%% file: main.tex
\documentclass[10pt,twocolumn,letterpaper]{article}

\usepackage[pagenumbers]{cvpr} 

\input{preamble}

\definecolor{cvprblue}{rgb}{0.21,0.49,0.74}
\usepackage[pagebackref,breaklinks,colorlinks,allcolors=cvprblue]{hyperref}

\def\paperID{} 
\def\confName{CVPR}
\def\confYear{2026}

\title{\titletext}

\author{
   Ellington Kirby\textsuperscript{1,*} \quad
   Alexandre Boulch\textsuperscript{1,*} \quad
   Yihong Xu\textsuperscript{1,*} \quad
   Yuan Yin\textsuperscript{1} \quad
   Gilles Puy\textsuperscript{1} \\
   Éloi Zablocki\textsuperscript{1} \quad
   Andrei Bursuc\textsuperscript{1} \quad
   Spyros Gidaris\textsuperscript{1} \quad
   Renaud Marlet\textsuperscript{1,2} \quad
   Florent Bartoccioni\textsuperscript{1} \\
   Anh-Quan Cao\textsuperscript{1} \quad
   Nermin Samet\textsuperscript{1} \quad
   Tuan-Hung VU\textsuperscript{1} \quad
   Matthieu Cord\textsuperscript{1,3}
   \vspace{0.3cm} \\
   \textsuperscript{1}valeo.ai, Paris, France \\
   \textsuperscript{2}LIGM, ENPC, IP Paris, UGE, CNRS, France \\
   \textsuperscript{3}Sorbonne Université, CNRS, ISIR, F-75005 Paris, France \\
   {\tt\small \{firstname.lastname\}@valeo.com}
}

\begin{document}

\maketitle
\begingroup
\renewcommand\thefootnote{*}
\footnotetext{Equal contribution}
\endgroup
\input{sec/0_abstract}    
\input{sec/1_intro}
\input{sec/2_related_work}
\input{sec/3_method}
\input{sec/4_x_ablations}
\input{sec/5_conclusion}
\section*{Acknowledgments}
{We thank Loick Chambon for constant support throughout the project and Lan Feng for helpful discussions. This work was granted access to the HPC resources of IDRIS under the allocations AD011016241, AD011016239R1 and AD011012883R4 made by GENCI. We acknowledge EuroHPC Joint Undertaking for awarding the project ID EHPC-REG-2024R02-210 access to Karolina, Czech Republic. 
Funded by the European Union. Views and opinions expressed are however those of
the author(s) only and do not necessarily reflect those of the European Union or the
European Commission. Neither the European Union nor the European Commission
can be held responsible for them. This work was supported by the European Union’s
Horizon Europe research and innovation programme under grant agreement No
101214398 (ELLIOT).
}

{
    \small
    \bibliographystyle{ieeenat_fullname}
    \bibliography{biblio}
}

\input{sec/X_suppl}


\end{document}

%% file: preamble.tex
\usepackage{fixmetodonotes}

\usepackage{graphicx}
\usepackage[dvipsnames]{xcolor}
\usepackage{colortbl}
\usepackage{pifont}

\usepackage{subcaption}
\usepackage{multirow}
\usepackage{wrapfig} 

\newif\ifshowedits

\newcommand{\addeditor}[3]{%
  \definecolor{#1color}{rgb}{#3}
  \expandafter\newcommand\csname #1\endcsname[1]{%
  \ifshowedits
    {\color{#1color} ##1}%
  \else
    {##1}%
  \fi
  }%
  \expandafter\newcommand\csname #1rmk\endcsname[1]{%
  \ifshowedits
    {\color{#1color} {\bf [#2: ##1]}}
  \fi
  }%
  \expandafter\newcommand\csname #1rpl\endcsname[2]{%
  \ifshowedits
    {\color{#1color} ##1 \sout{##2}}
  \else
    {##1}
  \fi
  }%
}

\newcommand{\titletext}{Driving on Registers}
\newcommand{\method}{DrivoR\xspace}

\definecolor{MyGreen}{RGB}{0, 180, 0}
\definecolor{MyRed}{RGB}{180, 0, 0}
\definecolor{MyYellow}{RGB}{180, 180, 0}
\newcommand{\cmark}{{\textcolor{MyGreen}{\ding{51}}}}%
\newcommand{\xmark}{{\textcolor{MyRed}{\ding{55}}}}%
\addeditor{alex}{AB}{0.5, 0.3, 0.8}
\addeditor{andrei}{Andrei}{0, 0.5, 0.5}
\addeditor{ek}{EK}{.1,.7,.3}
\addeditor{eloi}{EZ}{0.2, 0.2, .75}
\addeditor{yuan}{YY}{0.6, 0.6, 0.0}
\addeditor{gilles}{GP}{0.0, 0.45, 0.60}
\addeditor{renaud}{RM}{1, 0.4, 0.0}

\addeditor{yihong}{YH}{0.2, 0.2, 1.0}

\showeditstrue

\def\sut{s.t\onedot}

\newcolumntype{b}{>{\columncolor{blue!15}}c}

\usepackage{stmaryrd}
\usepackage{trimclip}

\makeatletter
\DeclareRobustCommand{\shortto}{%
  \mathrel{\mathpalette\short@to\relax}%
}

\newcommand{\short@to}[2]{%
  \mkern2mu
  \clipbox{{.5\width} 0 0 0}{$\m@th#1\vphantom{+}{\shortrightarrow}$}%
  }
\makeatother

%% file: sec/0_abstract.tex
\begin{abstract}
We present \method, a simple and efficient transformer-based architecture for end-to-end autonomous driving. Our approach builds on pretrained Vision Transformers (ViTs) and introduces camera-aware register tokens that compress multi-camera features into a compact scene representation, significantly reducing downstream computation without sacrificing accuracy. These tokens drive two lightweight transformer decoders that generate and then score candidate trajectories. The scoring decoder learns to mimic an oracle and predicts interpretable sub-scores representing aspects such as safety, comfort, and efficiency, enabling behavior-conditioned driving at inference. Despite its minimal design, \method outperforms or matches strong contemporary baselines across NAVSIM-v1, NAVSIM-v2, and the photorealistic closed-loop HUGSIM benchmark. Our results show that a pure-transformer architecture, combined with targeted token compression, is sufficient for accurate, efficient, and adaptive end-to-end driving. Code and checkpoints will be made available \href{https://valeoai.github.io/driving-on-registers/}{via the project page}.
\end{abstract}

%% file: sec/1_intro.tex
\section{Introduction}
\label{sec:intro}

\input{figures/architecture/architecture_tex}
End-to-end (E2E) planning has emerged as a promising direction 
for autonomous driving (AD), offering a single pipeline that maps raw sensor data and ego state to driving decisions \citep{hu2023uniad, paradrive}. Besides, by avoiding intermediate annotations such as 3D boxes, these methods reduce labeling cost. Among E2E approaches, trajectory-proposal methods, whether using a large pre-computed vocabulary \citep{chen2024vadv2,li2024hydra,li2025hydramdppp,li2025generalized}, or generating proposals on the fly \citep{liao2025diffusiondrive, xing2025goalflow, guo2025ipad, feng2025rap}, have shown particularly strong performance.

Methods 
predicting
multiple possible trajectories and selecting between them naturally capture the uncertainty within navigation. As in model-based RL \cite{moerland2023model}, the ability to score becomes central: the scorer must reliably choose the best candidate using context encoded in the sensor features.

The sensor processing backbone 
producing
the features that capture this context are thus key in E2E planning methods. These backbones typically dominate the parameter and FLOP count of E2E methods, often leveraging convolutional architectures like VoV-Net \citep{lee2019energy} or large pre-trained networks such as Vision Transformers \citep{Dosovitskiy2021vit} like EVA \citep{fang2023eva}, or DINO \citep{oquab2024dinov2}. Such backbones output thousands of tokens per frame, which must be processed for hundreds of trajectories. This creates a major computational bottleneck that only worsens as resolution or sensor count increases.

The most common solution to reducing the bottleneck is to pool these features along the spatial dimensions. However, feature pooling enforces specific resolution requirements on sensor inputs, and treats all inputs as equally informative, performing the same averaging operation across all cameras.
Inspired by works like \cite{yu2024image}, we ask just how many tokens are needed to represent a driving scene? 

We introduce \method, a ViT-based E2E planning architecture that replaces uniform pooling with a fixed set of per-camera register tokens that serve as compact scene descriptors. These tokens preserve planning-relevant context while drastically reducing the visual representation length.
Using this compressed representation, \method generates and scores trajectory proposals using two disentangled modules. The final trajectory is selected using predicted sub-scores, allowing behavior modulation at inference. 

Our method relies only on scoring annotations (rather than explicit 3D supervision) and achieves state-of-the-art results 
on NAVSIM-v1 \citep{dauner2024navsim}, NAVSIM-v2 \citep{Cao2025navsim2}, and the closed-loop HUGSIM benchmark \citep{zhou2024hugsim}.
Overall, our contributions are as follows:
\begin{itemize}
    \item An intentionally simple transformer architecture, without any intermediate BEV representations, or any large trajectory dictionary.
    \item The first work to explore specific, structural, advantages of ViT-based image backbones for E2E planning in the usage of register-based token compression.
    \item A disentangled scoring module enabling stronger performance and controllable behavior.
\end{itemize}

%% file: figures/architecture/architecture_tex.tex
\begin{figure*}[t]
    \centering
    \includegraphics[width=\linewidth]{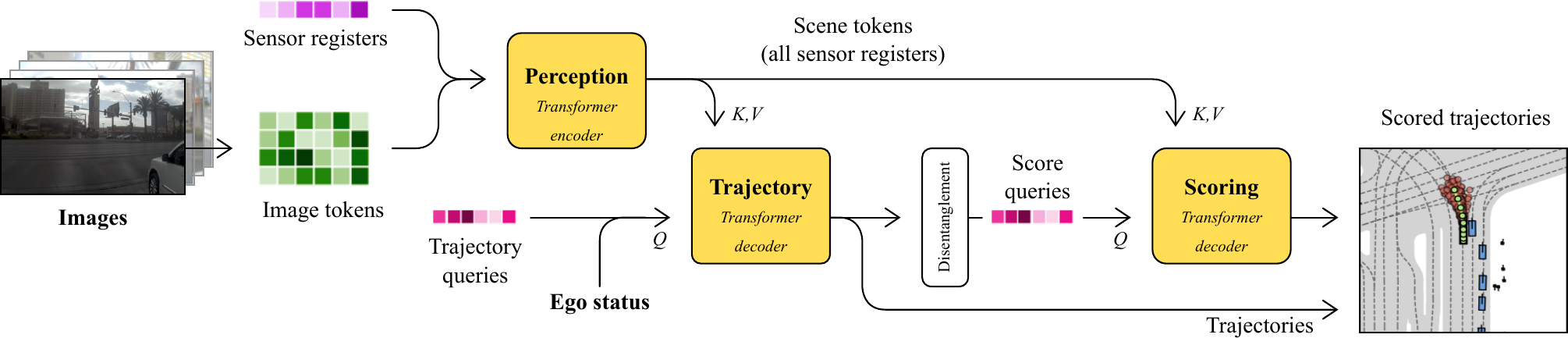}
    \caption{\textbf{DrivoR architecture.} The proposed architecture is composed of three transformer blocks: one encoder (perception) and two decoders (trajectory and scoring). The perception encoder compresses perceptual information in camera-aware registers for lightweight subsequent processing in the trajectory and scoring decoders. The decoded trajectories are re-embedded and detached from the gradient computation graph to disentangle scoring and generation. The final trajectory is chosen from the proposal set via the max predicted score.
    }
    \label{fig:architecture}
\end{figure*}


%% file: sec/2_related_work.tex
\section{Related Work}
\label{sec:related}

\paragraph{Token compression.}
Token reduction is central to efficient ViTs, whose attention cost grows quadratically with sequence length. Simple solutions such as patch-group pooling are parameter-free but treat all tokens uniformly. Other training-free strategies include matching-based compression \cite{bolya2023token}. Learned approaches range from Perceiver-IO’s latent queries \cite{jaegle2022perceiverio} to ViT register tokens, originally introduced to fix attention sinks \cite{darcet2023vision} and later used in compact generative models like TiTok \cite{yu2024image}. Recent driving-focused works \cite{ivanovic2025efficient,wang2025alpamayo} highlight the growing need for token reduction in real-time systems. To our knowledge, we are the first to repurpose ViT register tokens specifically for reducing visual tokens in E2E planning, enabling compact scene representations while retaining planning-critical context.




\paragraph{End-to-end driving.}
End-to-end learning has become popular in autonomous driving since the proof of concept of pioneering works \cite{hu2023uniad, jiang2023vad} like UniAD \cite{hu2023uniad}. However, they still heavily rely on modular designs with different sub modules, such as detection, tracking and mapping, making it hard to deploy. With the introduction of more efficient (pseudo) closed-loop evaluation metrics~\cite{dauner2024navsim, Cao2025navsim2} or simulation~\cite{zhou2024hugsim}, more recent methods~\cite{guo2025ipad, liao2025diffusiondrive, chitta2022transfuser, li2025recogdrive, chen2024vadv2} predict the planned trajectory or actions directly from sensor inputs, a step forward toward the fully end-to-end paradigm.

These E2E methods \cite{guo2025ipad,liao2025diffusiondrive, li2025recogdrive} mostly rely on off-the-shelf CNN-based (e.g., ResNet-34, ResNet-50, V2-99) image encoders without putting further attention on the design of the perception stack. Downstream in the planning stack, transformers are more robustly explored. Transfuser \cite{chitta2022transfuser} uses transformers for LiDAR–image fusion; DriveTransformer \cite{jia2025drivetransformer} unifies multiple tasks under one transformer; and iPad \cite{guo2025ipad} exploits transformer residuals for iterative trajectory refinement. Innovation has concentrated on how to use mid-level sensor features rather than on stronger pretrained backbones. When ViTs~\cite{Dosovitskiy2021vit} are used, methods~\cite{feng2025rap, li2025generalized, yao2025drivesuprim, li2025ztrs} typically rely on large or huge variants without considering the computation costs. These models lean on attention to fuse heterogeneous inputs (ego state, images, LiDAR, commands), but often supplement it with costly 3D reasoning-BEV projections, deformable cross-attention, or LiDAR supervision, raising annotation and compute demands. In this context, we design a simple query-based transformer architecture, built on the much smaller ViT variants, avoiding the high cost of complex intermediate representation, while keeping superior E2E driving performance.

\paragraph{Trajectory scoring.} 
Trajectory-proposal based planning has become a compelling direction for E2E driving, introduced in Hydra-MDP \cite{li2025hydramdppp}. Producing many possible futures and selecting one forces the model to address multimodality head-on. This shifts importance toward the scoring module, which must reliably choose the best candidate \cite{liao2025diffusiondrive}. A recent state-of-the-art work on NAVSIM-v2, GTRS \cite{li2025generalized} showed that learning a sufficiently strong scorer using a ViT backbone, paired with a large trajectory vocabulary, can solve complex scenarios. We introduce the importance of separate scoring and trajectory generation pathways in our E2E model.

%% file: sec/3_method.tex
\section{Method}
\label{sec:method}

\subsection{A Simple Design}

We design a simple and efficient transformer based architecture for planning in autonomous driving.
The overall pipeline is presented in \autoref{fig:architecture}.
It is composed of three modules: an encoder for perception and two decoders for trajectory estimation and scoring. The model follows a classical transformer encoder-decoder architecture~\cite{vaswani2017attention} without any complex intermediate representation.

The \emph{encoder for perception} (see~\autoref{sec:method_perception}) is a vision transformer~\cite{Dosovitskiy2021vit} applied per camera.
In order to compress the feature map to small set of tokens, we introduce additional per-camera registers, finetuned with the backbone to encode the perceptual information. We group together all per-camera register tokens at the output of the ViT to form the scene tokens.

The \emph{trajectories} are then estimated from learned queries using a transformer decoder attending to the scene tokens via cross-attention (see \autoref{sec:method_decoding_trajectories}).

Finally, the trajectories are scored in the \emph{scoring decoder}. Each trajectory becomes a query input of the decoder, which also attends to the scene tokens to produce scores (see \autoref{sec:method_decoding_scoring}).

During training, the trajectories are learned using a winner-takes-all regression loss (\autoref{sec:method_trajectory_loss}) and the scores are learned against an oracle scorer, e.g., provided along with the dataset (\autoref{sec:method_score_loss}). At inference, we re-interpret our learned scoring function as a reward function, enabling  driving according to several behavior conditioned policies with a single trained model.

\subsection{Perception Encoder}
\label{sec:method_perception}
\input{figures/architecture/architecture_encoder_decoder}

At perception level, we seek to fulfill three objectives: 1) leveraging recent ViT architectures; 2) benefiting from pretrained weights; and 3) limiting the output size to contain the computational complexity in the decoders.
To this end, we efficiently compress the perceptual signals into a limited set of tokens using additional registers, and finetune the backbone with LoRA. as illustrated in \autoref{fig:architecture_encoder_decoder}a.

For each camera, we concatenate $R$ camera registers of size $D_\text{ViT}$, along with pre-existing registers, classification token, and patch tokens. $D_\text{ViT}$ denotes the dimension of the ViT features. All registers and tokens are fed to the ViT. We then retrieve the $R$ camera tokens at the final layer of the ViT. The final camera tokens for each of the $N$ cameras are finally grouped together to obtain $N \times R$ scene tokens of size $D_\text{ViT}$.
Note that we use per-camera registers, that is we initialize $N \times R$ registers, where $N$ is the number of input cameras.
This allows us to have camera-aware scene tokens: the model can differentiate if a given scene token is extracted from, e.g., the front, left or right camera.

This compression into a small set of camera tokens is close in spirit to Perceiver approaches \citep{jaegle2021perceiver,jaegle2022perceiverio}, with the noticeable difference that these approaches use cross-attentions for compression. 
Setting up such a mechanism in the 
encoder would require changes in the ViT architecture.

In our case, we can directly use a pretrained ViT as initialization, and
perform LoRA finetuning of the ViT backbone to learn the vision-to-register compression, reducing parameter count and speeding training. 

\subsection{Trajectories}
\label{sec:method_decoding_trajectories}
\paragraph{Trajectory decoder.}

All decoders use the architecture depicted in \autoref{fig:architecture_encoder_decoder}b,
which consists of a vanilla transformer decoder~\cite{vaswani2017attention}: a stack of $k$ transformer blocks, each made of a self-attention layer, followed by a cross-attention to the scene tokens, and a feed-forward network (FFN)
, all with residual connections.

The input of the trajectory decoder consists of a set of learnable trajectory queries $Q_\text{traj}$, each of dimension $D_\text{traj}$, which is also the inner dimension of the trajectory decoder. 
After the process, the queries are decoded into $\vert Q_\text{traj}\vert$ candidate trajectories.
These queries are randomly initialized and learned during training.
The ego status inputs, consisting of poses, velocities, accelerations, and driving command, are encoded and added to the trajectory queries before entering the decoder. 
The final trajectory tokens at the end of the transformer are decoded into trajectories with an MLP.

Each decoded candidate trajectory $\tau_i$ is a sequence of $n_p$ poses predicted from the current timestep $t$ (excluded) up to a future horizon at $t+T$, 
$T$ being
the total prediction duration. Each pose is represented as $(x, y, \theta)\in\mathbb{R}^3$, and the full trajectory lies in $\mathbb{R}^{n_p\times 3}$. The variables $x$ and $y$ denote the longitudinal and lateral displacements, respectively, and $\theta$ is the heading. All quantities are expressed in the ego agent's local reference frame at timestep $t$. The time interval between successive predicted poses is assumed to be uniform.

\paragraph{Trajectory loss.}
\label{sec:method_trajectory_loss}
The trajectories ${\tau_i}$ are learned using a Winner-Takes-All (WTA) approach~\cite{guzman2012multiple} or, equivalently, the minimum-over-n (MoN) loss~\cite{guo2025ipad}. Given a reference human trajectory $\hat{\tau}$ of duration $T$ and consisting of $n_p$ poses, only the closest predicted trajectory is supervised, which allows the model to produce diverse candidate trajectories:
\begin{equation}
    \mathcal{L}_\text{traj} = \min_i \left\lVert \tau_i - \hat\tau \right\rVert_1
\end{equation}
This formulation encourages the model to consider multiple plausible pathways for a given scene. 

An additional regression target $\hat{\tau}'$ can be introduced to encourage the predicted trajectories to reach farther waypoints. To construct this target, a reference trajectory with duration $T' > T$ is resampled to $T$ using cubic spline interpolation, producing an accelerated version that matches the number of predicted poses $n_p$. The resulting multi-target trajectory loss is written as
\begin{equation}
    \mathcal{L}_\text{traj} = \min_i (\left\lVert \tau_i - \hat\tau \right\rVert_1 + \left\lVert \tau_i - \hat\tau' \right\rVert_1)\>.
\end{equation}

\subsection{Scores}
\paragraph{Scoring decoder.}
\label{sec:method_decoding_scoring}
The scoring decoder, which evaluates the quality of each candidate trajectory, uses an architecture mirroring that of trajectory generation. 

The scoring decoder takes as input the decoded trajectories as well as perceptual information through the scene tokens. Each decoded trajectory is turned into a $D_\text{score}$-dimensional query using an MLP. All the trajectory queries are fed to the scoring decoder.

Embedding the decoded trajectories into a new feature space rather than reusing the trajectory decoder's output tokens is key in our architecture. This enforces a separation between the information used to \emph{generate} trajectories and the information used to \emph{score} them: the scorer sees only the decoded trajectory, not the additional latent details still present in the trajectory tokens.

Our scoring decoder then uses cross-attentions between scene tokens and score queries, allowing gradients to flow back to the perception encoder to learn scene tokens useful for both trajectory and scoring predictions. However, we prevent the gradient from the scoring decoder from flowing back to the trajectory decoder. This prevents the trajectory decoder from being influenced by the current quality of the scoring decoder during training.

Finally we predict the six score components \cite{dauner2024navsim} using a dedicated MLP for each score.

\paragraph{Scoring for adjustable driving behavior.}

A key feature of our model is its ability to adapt trajectory selection to different driving preferences. For instance, one user may prioritize safety and comfort, while another may favor faster progress at the cost of smoothness. To enable such flexibility, our scoring head predicts separate sub-scores corresponding to different aspects of driving quality (e.g., safety, comfort, efficiency). These sub-scores can then be combined post-hoc into a single meta-score at inference time, allowing the relative importance of each term to be adjusted without retraining the model. We adopt the sub-scores directly from the Predictive Driver Model Score (PDMS) scorer used in NAVSIM-v1. 

\paragraph{Scoring loss.}
\label{sec:method_score_loss}
We train our scoring network via binary cross entropy (BCE) to predict the individual sub-score components $c$ of the PDMS, each given a weight $\lambda_c$:
\begin{equation}
\label{eq:score}
    \mathcal{L}_\text{score} =  \sum_{c}  \lambda_c  \sum_i \operatorname{BCE}\left(\mathcal{G}_{\theta_c}(\tau_i), \mathcal{G}_c(\tau_i) \right)
\end{equation}
where $\mathcal{G}_c$ is an oracle scorer for the sub-score $c$, and $\mathcal{G}_{\theta_c}(\tau_i)$ is our learned scoring head for sub-score $c$.

\paragraph{Inference.}
At inference, we interpret our scoring network as a reward function and our full pipeline as a driving policy which is conditioned on a specific behavior profile encapsulated in the set of weights applied to the scoring outputs. Borrowing from Offline-RL literature such as CtRL-Sim \cite{rowe2024ctrl}, we can thus condition our final scoring output on a driving behavior by modifying the values of our $\lambda_c$,
and selecting trajectories which maximize the score computed with this new set of weights. For example, this can encourage trajectories which make maximal estimated progress.

\subsection{Final Training Loss.}
The training loss is the combination the trajectory loss and the scoring loss:
\begin{equation}
    \mathcal{L} = \mathcal{L}_\text{traj} + \lambda_s \mathcal{L}_\text{score}.
\end{equation}
In practice during training, the weights of the losses are all set to 1 for ease of implementation, \sut, $\lambda_c = 1$ for each sub-score $c$ and $\lambda_s=1$.

%% file: figures/architecture/architecture_encoder_decoder.tex
\begin{figure}[t]
    \centering
    \small
    \setlength{\tabcolsep}{7pt}
    \begin{tabular}{@{}c|c@{}}
    \includegraphics[width=0.42\linewidth]{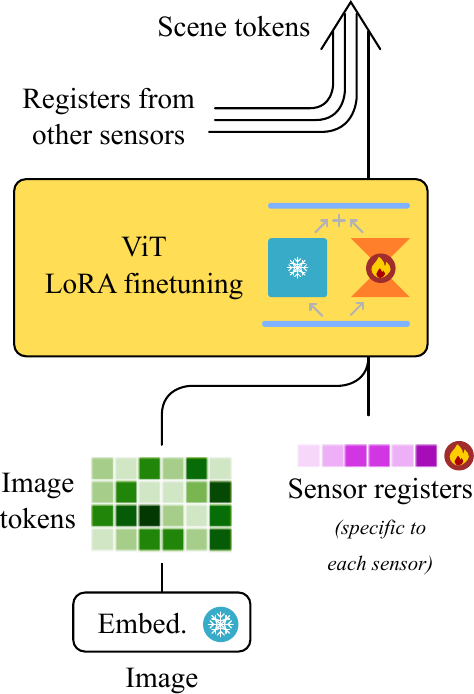}&
    \includegraphics[width=0.48\linewidth]{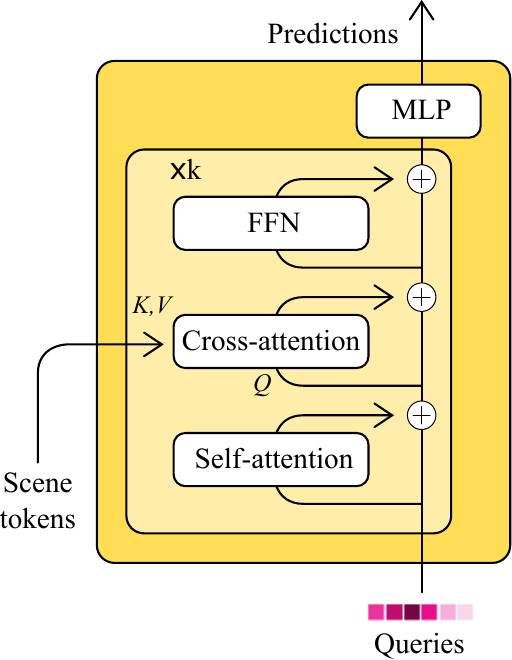} \\
    (a) Encoder & (b) Decoder \\
    \end{tabular}
    \caption{\textbf{Encoder and decoder architectures} follow standard transformer architectures, with introduction of sensor registers in the encoder, and using these registers as scene tokens in downstream decoders.}
    \label{fig:architecture_encoder_decoder}
\end{figure}

%% file: sec/4_x_ablations.tex
\section{Experiments}
\label{sec:ablations}
\paragraph{Experimental setup.}
As input, we use 4 cameras (front, front left, front right and back).
The perception module is a DINOv2 ViT-S \citep{oquab2024dinov2} encoder, LoRA finetuned \citep{hu2022lora} (rank 32) following~\cite{barin2024robust}.
By default, we add 16 registers per camera. 
The decoders are 4-layer transformers with an inner dimension of 256.
The feed-
forward network has a dilation factor of 4.
The camera registers as well as the initial trajectory tokens are randomly initialized with normal distribution $\mathcal{N}(0,10^{-6})$.
The model is trained on the \texttt{navtrain} split for 10 epochs, with learning rate $2\times 10^{-4}$ and cosine annealing scheduling.
For the ablation studies, if not described otherwise, we train our model with this common base architecture. All ablation model scores are computed on the \texttt{navval} split. All models were trained on 4 NVIDIA A100 GPUs. The \method model is roughly \textbf{40M} parameters, significantly below comparable works.

\subsection{Benchmarks}

\paragraph{NAVSIM-v1.}
NAVSIM-v1~\cite{dauner2024navsim} is a dataset built out of nuPlan~\cite{caesar2021nuplan} as a subset of OpenScene~\cite{contributors2023openscene}.
As opposed to previous benchmark such as nuScenes~\cite{nuscenes}, mostly formulating driving quality as a measure of similarity to the expert human trajectory, NAVSIM-v1 introduces metrics inspired from closed-loop simulation.
The main metric, Predictive Driver Model Score (PDMS), is an aggregation of penalties, e.g., collisions with other, non-reactive, agents, staying on-road and in the correct direction, and avoiding near-misses with other agents. These scores are combined with quality-related scores such as the comfort and progress. Progress is measured as a comparison to the centerline-progress of a PDM agent \cite{dauner2023pdm} given access to complete ground-truth information. For training, we use default PDMS weights defined in the benchmark~\cite{dauner2024navsim}. 

The results obtained on the NAVSIM-v1 benchmark are presented in \autoref{tab:benchmark_navsim_v1_small}.
\method outperforms all other methods on NAVSIM-v1, and nears human-level performance. We highlight the comparison to RAP \cite{feng2025rap}, which we consider an orthogonal work: the large quantity of rasterized data introduced in the paper
may be used in any method.

\input{tables/navsim_v1_small}
\input{tables/hugsim_small}

\input{tables/navsim_v2_small}

\paragraph{NAVSIM-v2.}
NAVSIM-v2~\cite{Cao2025navsim2} builds on NAVSIM-v1 with the objective of closing the gap with closed-loop, simulator driven, benchmarks. NAVSIM-v2 introduces a second stage of evaluation, where novel variations of scenes are generated via Gaussian Splatting. These novel scenes consist of perturbations of the ego vehicle status, i.e., shifts and rotations, forcing the model to generalize outside of the training distribution. NAVSIM-v2 is scored using an extended version of the score from NAVSIM-v1, termed the EPDMS. We present the results on the \texttt{navhard-two-stage} split of NAVSIM-v2 in table \autoref{tab:benchmark_navsim_v2_small}, where \method outperforms all existing works. We note that the results in \autoref{tab:benchmark_navsim_v2_small} were computed after an official bug fix, and thus do not include the largest GTRS-Dense model. Results on NAVSIM-2 before the fix are included in the supplementary material. 
We use \texttt{warmup-two-stage}\footnote[1]{\texttt{warmup-two-stage} intersects with \texttt{navhard-two-stage}, after request, benchmark authors validated its use as validation set.
} as validation set for NAVSIM-v2 containing 7 scenes.

We highlight GTRS-DrivoR-ViT-S (row 4), produced by replacing the GTRS backbone with our ViT-S + register-based compression while keeping their original vocabulary and scorer. This swap removes GTRS's pooling in favor of our learned compression, improving performance over the similarly sized V2-99 backbone and approaching the EVA-ViT-L variants (see supplementary), while delivering over 3× higher throughput. Notably, our full SOTA model differs from this variant only in the scoring pipeline, demonstrating the gains from our scorer.

\paragraph{HUGSIM.}
For the closed-loop evaluation, we use HUGSIM~\cite{zhou2024hugsim}, a benchmark with scenarios adapted from the original scenes of KITTI-360~\cite{kitti360}, nuScenes~\cite{nuscenes}, PandaSet~\cite{pandaset}, and Waymo~\cite{waymo}. These scenarios are reconstructed as photorealistic 3D environments in which an E2E model controls the ego agent to navigate within each scenario. The planner perceives the 3D environment through RGB cameras whose viewpoints are determined by the ego agent's position in the scenario.


\autoref{tab:hugsim} reports results on the pre-challenge HUGSIM test set (345 scenarios across four difficulty levels). We follow the zero-shot protocol and evaluate with Road Completion (RC) and the HUGSIM Driving Score (HD-Score), the latter combining RC with averaged NC, DAC, TTC, and comfort. DrivoR, trained only on NAVSIM-v1 with no finetuning, achieves an RC of 49.8 and an HD-Score of 35.7—the highest among the reported baselines.

Note that we identified several anomalies in the official evaluation and simulation code, including inconsistent acceleration bounds in the comfort metric computation and an incorrect heading computation for the planned trajectory provided to the controller. After applying the necessary fixes, we reproduced all scores using our corrected implementation, since the results produced by the original code are not directly comparable and do not ensure a reliable evaluation.

\paragraph{Efficiency.} 
We benchmark the runtime performance of \method against a ViT-L baseline (GTRS) on a single element batch and an A100 GPU without quantization nor acceleration process. \method obtains a more than 3x throughput improvement (from 400ms/forward to 110ms/forward), with 3x reduction in GFLOPS and peak memory usage, with improved scores, demonstrating a step towards real time usage of a ViT backbone.

\begin{figure}[h] 
    \centering
    \begin{subfigure}[b]{0.45\textwidth}
        \centering
        \includegraphics[trim={0cm 0cm 0cm 0cm}, clip, width=\linewidth]{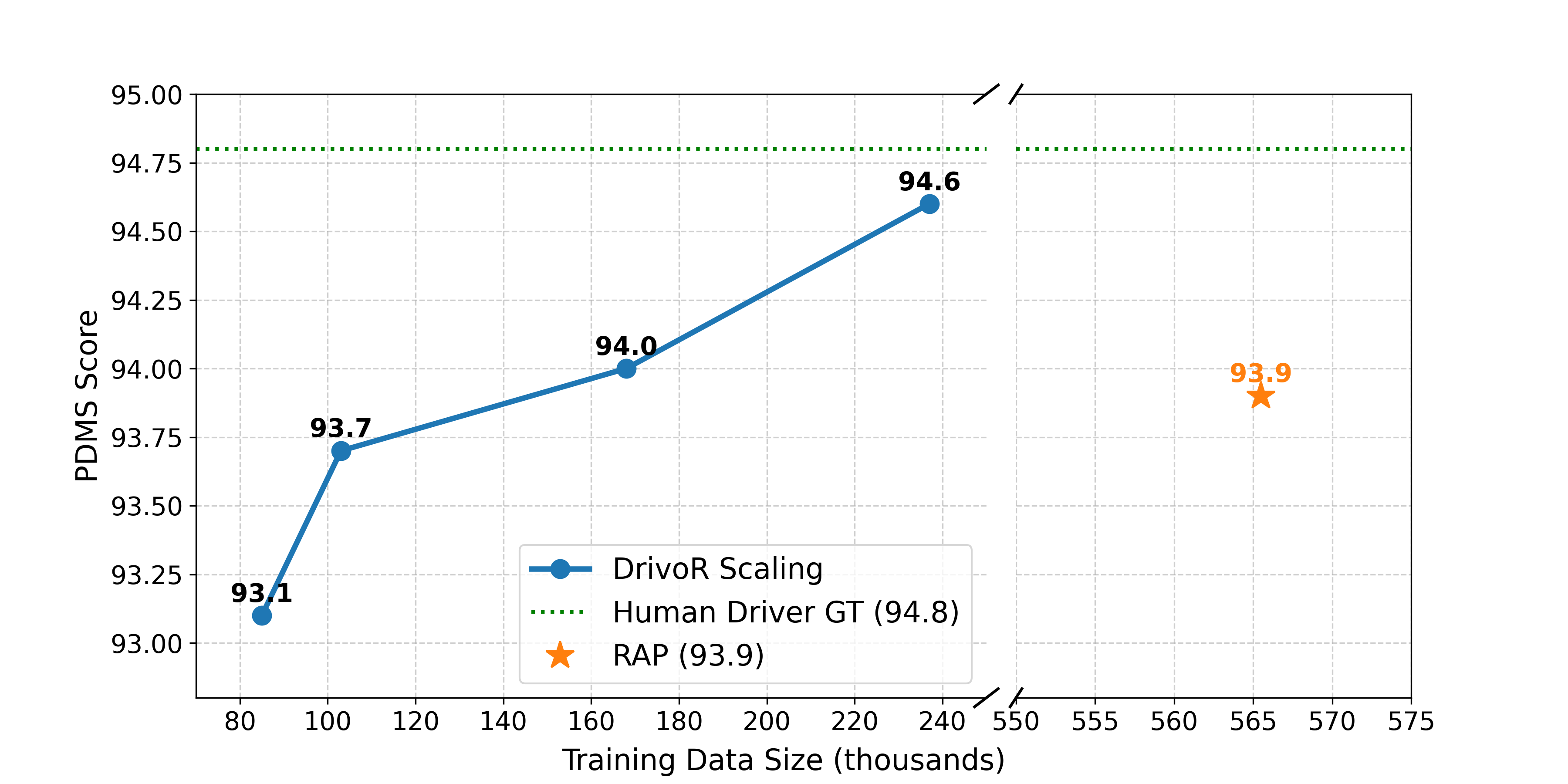}
        \caption{NAVSIM-v1 (PDMS)}
        \label{fig:nav1_results}
    \end{subfigure}
    \hfill 
    \begin{subfigure}[b]{0.45\textwidth}
        \centering
        \includegraphics[trim={0cm 0cm 0cm 0cm}, clip, width=\linewidth]{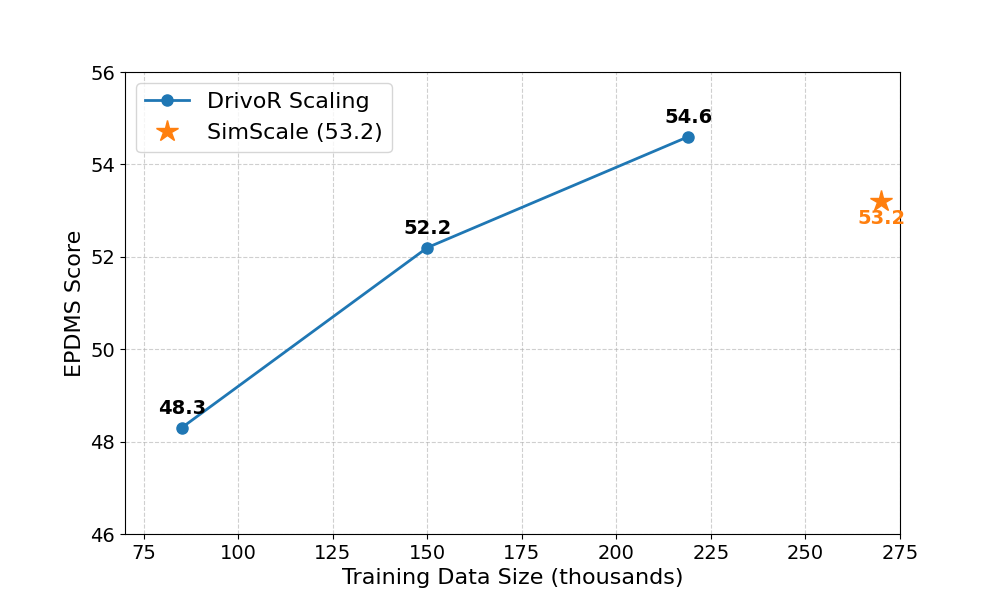}
        \caption{NAVSIM-v2 (EPDMS)}
        \label{fig:nav2_results}
    \end{subfigure}
    \caption{\textbf{\method Scalability.} Scaling \textbf{DrivoR} (blue) against current SOTA baselines (orange). (a) On the NAVSIM-v1 task, DrivoR surpasses both the RAP~\cite{feng2025rap} VIT-H baseline. (b) On the NAVSIM-v2 task, DrivoR achieves a new state-of-the-art score of 54.6 EPDMS.}
    \label{fig:scaling_results}
\end{figure}

\paragraph{Scaling \method.}
{With its simple design using vanilla transformers and registers, we study the scalability of \method. We leverage synthetic data from \cite{tian2025simscale} with pseudo annotations from PDM-Closed~\cite{dauner2023pdm}. Concretely, we extract 65k and 134k annotations from \cite{tian2025simscale} and jointly train with \texttt{navtrain}, respectively. We show in \autoref{fig:scaling_results} that \method consistently improves the NAVSIM performance with the increased data. Moreover, \method sets a new state-of-the-art performance with less data, compared to RAP~\cite{feng2025rap} using 500k+ synthetic data and SimScale~\cite{tian2025simscale} with 185k extra data. More detailed performance scores are provided in \autoref{tab:benchmark_navsim_v1_small} and \autoref{tab:benchmark_navsim_v2_small}.}

\subsection{Ablation Studies}
\subsubsection{Perception}
\input{tables/ablations_perception_all}

\paragraph{Image backbone initialization.}
Various previous methods~\cite{feng2025rap, guo2025ipad} use pretrained backbones.
We first study the impact of such an initialization on the planning results in \autoref{tab:ablations_pretraining}.
First, a good initialization of the perception ViT is crucial: using a pretrained backbone improves scores by a large margin (more than 15 PDMS points).
Second, using large-scale pretrained DINOv2~\cite{oquab2024dinov2} improves further over pretraining on ImageNet21k~\cite{kolesnikov2020big}.
In all subsequent experiments, we use a pretrained DINOv2 backbone.

\paragraph{Compression to low number of perceptual tokens.}
As argued in \autoref{sec:method}, using few perceptual tokens is of interest as it makes the trajectory prediction and scoring lightweight and faster.
In \autoref{tab:ablations_perception_backbone}, first column, we study three compression approaches: 
using a pooling operation of the output feature map as in \cite{li2025generalized} (c), using a transformer decoder with 16 queries per camera (e), and our proposed approach using 16 additional registers in the model (h).
For comparison purpose, we also provide models using the full feature maps, i.e., 16k scene tokens in (b).
We observe that our register-based approach outperforms the pooling operation while introducing a low-overhead (0.6M parameters) at training.
It also nearly reaches the performances of the no-compression model, despite using 250 times fewer tokens for downstream processing.
Additionally, we observe that using a transformer decoder (with roughly the same number of parameters) does not reach the same performance.

\paragraph{Register token correlation.}

\autoref{fig:token_qualitative} shows inter-register cosine similarity for each camera.
Front-camera tokens are largely de-correlated, suggesting per-register specialization.
\autoref{fig:token_attention} confirms this: different front camera registers attend to distinct regions (e.g., lead vehicle, traffic lights, sidewalk).

Moving toward the rear cameras, similarity increases sharply: most tokens collapse to the same representation, and within the back camera only one remains distinct. The attention maps (second row of \autoref{fig:token_attention}) reflect this collapse.

This pattern aligns with driving intuition: most attention is devoted to the scene ahead, with only brief checks behind. We hypothesize that collapsing less informative views (side/back) reduces noise for downstream planning, an effect which is impossible to observe with uniform pooling.
\input{figures/token_qualitative/token_qualitative}

\input{figures/token_qualitative/token_attention_maps}

\paragraph{Finetuning strategy.}

Next, we study the finetuning strategy in \autoref{tab:ablations_perception_backbone}, second column, with three different settings: full finetuning (d, f), frozen backbone (a, d, g) and LoRA finetuning~\cite{hu2022lora} (b, c, e, h).
For frozen backbone with registers, only the registers are learnable.
First, we observe that LoRA finetuning improves the results by a large margin, reaching similar conclusion as in \citep{barin2024robust} over frozen backbone.
It is also the case when comparing LoRA (h) and Full finetuning (f), but to our understanding, it should be possible to close this gap with careful learning rate scheduling specific to the backbone.
LoRA being more robust to these meta parameters, we use LoRA finetuning as default.

\paragraph{Number of registers.}

We present in \autoref{tab:ablations_num_tokens} the evolution of the validation score depending on the number of camera tokens.
We either use new registers randomly initialized (the 4 original registers of the DINOv2 backbone are discarded) or, we keep the DINOv2 registers along with our randomly initialized ones.
In the later case, to discriminate the cameras, we add per-camera learnable positional encoding to the DINOv2 registers at the image backbone output.

First, using the DINOv2 registers does not help compared to random initialization. We observe that the already specialized registers could be a bad initialization for driving tasks. We hypothesize that similarly to full finetuning, a careful learning rate scheduling could mitigate this gap.

Second, we observe that using more registers improves the performances up to a plateau between 16 and 32 registers per camera, we thus select 16 registers.

\subsubsection{Trajectories}

The number of trajectory queries strongly affects performance. As shown in \autoref{tab:ablations_scaling_trajectories}, increasing queries from 1 (human-only regression) to 128 yields gains that plateau around 64, which we adopt as our default.
\input{tables/ablations_trajectory}




\input{tables/ablations_scoring_tiny}

\subsubsection{Scoring}
\autoref{tab:ablations_scoring} shows our scoring pipeline ablation results.

\textbf{Single vs.\ dual branches:} (a) Using one transformer branch for both trajectory generation and scoring (with separate MLP heads) harms performance. As shown in \autoref{fig:scoring_qualitative}, the two tasks attend to different cameras: generation focuses on the front view even for left turns, while scoring draws on left- or rear-camera features depending on trajectory sharpness or collision risk. This motivates separate branches. 

\textbf{Disentanglement:} We next examine whether the scoring branch should embed trajectories in a new space and block gradients to the generator. Variant (b) does not disentangle; (c, d) do. Increased separation consistently improves performance. 

\textbf{Sub-score prediction:} Comparing (c) and (d), predicting multiple sub-scores not only enables behavior control but also improves accuracy. This is likely due to the fact that learning the final output of the scoring formula in \autoref{eq:score} is more difficult than predicting its components.

\input{figures/scoring_qualitative/scoring_qualitative}

\subsubsection{Training}

\paragraph{Longer trajectory.}
In \autoref{sec:method}, we introduced an augmentation that regresses to a second, more aggressive trajectory. \autoref{tab:ablations_loss_long} compares this double objective to standard regression on the human trajectory. It improves NAVSIM-v1, which rewards progress over comfort, but hurts performance on NAVSIM-v2 \texttt{warmup-two-stage}, where perturbed agent states require more cautious driving to avoid collisions and recover safely.
\input{tables/ablations_loss_long_traj}

\paragraph{Final training setup.}
For the final model setting, we evaluated longer training lengths.
We observe that PDMS on \texttt{navval} increase with the number of epochs, reaching a plateau at 25 epochs.
Our final model for NAVSIM-v1 is then trained for 25 epochs, on the competition split (\texttt{navtrain} + \texttt{navval}). On NAVSIM-v2 (\texttt{warmup-two-stage}), we observe an opposite trend, with EPDMS decreasing with the number of training epochs as well as training using the competition split. We therefore use 10 epochs for NAVSIM-v2.

\paragraph{Safety-oriented agent.}
We evaluate the behavior of our agent obtained after tuning the score weights on \texttt{warmup-two-stage}. Intuitively we expect safer driving to be required for NAVSIM-v2 due to out-of-distribution scenes. In \autoref{fig:safety_agent} we represent this agent in dark blue (``Safety-Oriented Agent''), and indeed see improved performance on safety and comfort metrics, but decreased progress, representing more passive, safe, driving.

\input{figures/safety_agent/safety_agent}

%% file: tables/navsim_v1_small.tex
\begin{table}[t]
    \centering
    \small
     \setlength{\tabcolsep}{1.8pt}
    \resizebox{\columnwidth}{!}{
    \begin{tabular}{@{}l@{}rl|ccccc|c@{}}
    \toprule
    Method & & & NC & DAC & TTC & Comf.& EP & \textbf{PDMS} \\
    \midrule
    \rowcolor{black!10}
    PDM‑Closed & \cite{dauner2023pdm}               & \scriptsize{PMLR'23} & 94.6 & 99.8 & 89.9 & 86.9 & 99.9 & 89.1 \\ 
    \rowcolor{black!10}
    Human driver& \cite{dauner2024navsim}           & \scriptsize{NeurIPS'24} & 100 & 100&  100 & 99.9 & 87.5 & 94.8 \\ 
    \rowcolor{black!10}
    RAP-DINO$^{\dagger}$  & \cite{feng2025rap}                   & \scriptsize{arXiv'25} & 99.1 & 98.9 & 96.7 & 100 & 90.3 & 93.8 \\ 
    \midrule
    Ego‑stat. MLP & \cite{dauner2024navsim}        & \scriptsize{NeurIPS'24} & 93.0 & 77.3 & 83.6 & 100 & 62.8 & 65.6 \\ 
    UniVLA  & \cite{wang2025unified}                & \scriptsize{arXiv'25} & 96.9 & 91.1 & 91.7 & 96.7 & 76.8 & 81.7 \\ 
    DrivingGPT & \cite{chen2024drivinggpt}          & \scriptsize{ICCV'24} & 98.9 & 90.7 & 94.9 & 95.6 & 79.7 & 82.4 \\ 
    UniAD & \cite{hu2023uniad}                   & \scriptsize{CVPR'23} & 97.8 & 91.9 & 92.9 & 100  & 78.8 & 83.4 \\ 
    LTF & \cite{chitta2022transfuser}               & \scriptsize{TPAMI'22} & 97.4 & 92.8 & 92.4 & 100  & 79.0 & 83.8 \\ 
    PARA‑Drive & \cite{paradrive}               & \scriptsize{CVPR'24} & 97.9 & 92.4 & 93.0 & 99.8 & 79.3 & 84.0 \\ 
    DriveX-S & \cite{shi2025drivex}                 & \scriptsize{ICCV'25} & 97.5 & 94.0 & 93.0 & 100  & 79.7 & 84.5 \\ 
    World4Drive & \cite{zheng2025world4drive}       & \scriptsize{ICCV'25} & 97.4 & 94.3 & 92.8 & 100  & 79.9 & 85.1 \\ 
    DRAMA & \cite{yuan2024drama}                    & \scriptsize{ISRR'24} & 98.0 & 93.1 & 94.8 & 100  & 80.1 & 85.5 \\ 
    VAD-v2 & \cite{chen2024vadv2}                   & \scriptsize{arXiv'24} & 98.1 & 94.8 & 94.3 & 100 & 80.6 & 86.2 \\ 
    PRIX & \cite{wozniak2025prix}                   & \scriptsize{arXiv'25} & 98.1 & 96.3 & 94.1 & 100  & 82.3 & 87.8 \\ 
    DiffusionDrive & \cite{liao2025diffusiondrive} & \scriptsize{CVPR'25} & 98.2 & 96.2 & 94.7 & 100  & 82.2 & 88.1 \\ 
    DIVER & \cite{song2025breaking}                 & \scriptsize{arXiv'25} & 98.5 & 96.5 & 94.9 & 100  & 82.6 & 88.3 \\ 
    AutoVLA & \cite{zhou2025autovla}                & \scriptsize{NeurIPS'25} & 98.4 & 95.6 & 98.0 & 99.9 & 81.9 & 89.1 \\ 
    DriveVLA-W0 & \cite{li2025drivevla}             & \scriptsize{arXiv'25} & 98.7 & 99.1 & 95.3 & 99.3 & 83.3 & 90.2 \\ 
    ReCogDrive & \cite{li2025recogdrive}            & \scriptsize{arXiv'25} & 97.9 & 97.3 & 94.9 & 100  & 87.3 & 90.8 \\ 
    Hydra-MDP++   & \cite{li2025hydramdppp}         & \scriptsize{arXiv'25} & 98.6 & 98.6 & 95.1 & 100  & 85.7 & 91.0 \\ 
    iPad & \cite{guo2025ipad}                       & \scriptsize{arXiv'25} & 98.6 & 98.3 & 94.9 & 100  & 88.0 & 91.7 \\ 
    Centaur & \cite{sima2025centaur}                & \scriptsize{arXiv'25} & 99.5 & 98.9 & 98.0 & 100  & 85.9 & 92.6 \\ 

    DriveSuprim  & \cite{yao2025drivesuprim}         & \scriptsize{arXiv'25} &98.6 & 98.6 & 95.5 & 100 & 91.3 & 93.5 \\
    \rowcolor{blue!15}
    \method{} \tiny{(train)}   &&& 98.9 & 98.3 & 96.2 & 100 & 89.1 &  93.1      \\
    \rowcolor{blue!15}
    \method{} \tiny{(trainval)}   &&& 99.0 & 98.9 & 96.7 & 100 & 90.0 &  93.7      \\
    \rowcolor{blue!15}
    \method{} \tiny{(+65k SimScale data)}   &&& 99.1 & 99.0 & 96.9 & 100 & 90.3 & 94.0      \\
    \rowcolor{blue!15}
    \method{} \tiny{(+134k SimScale data) \hspace{+1mm}}   &&& 99.1 & 99.2 & 96.9 & 100 & 91.6 & \bf 94.6      \\
    \bottomrule
    \multicolumn{9}{@{}l@{}}{$^\dagger$: \scriptsize{RAP~\cite{feng2025rap} is trained on a dataset that is 10$\times$ larger than navtrain (the default training set).}}
    \end{tabular}
    }
    \caption{\textbf{NAVSIM-v1.} Comparison to existing camera-only methods on the NAVSIM-v1 benchmark on test set (\texttt{navtest}). Full definition of scores in supplementary material, higher is better.
    }
    \label{tab:benchmark_navsim_v1_small}
\end{table}

%% file: tables/hugsim_small.tex
 \begin{table}[t]
    \centering
    \small
    \setlength{\tabcolsep}{1.5pt}
    \begin{tabular}{@{}l@{}r|ccccc|ccccc@{}}
    \toprule

    & & \multicolumn{5}{c}{RC} & \multicolumn{5}{c}{HD-Score} \\
    Method & & E & M & H & X & Avg. &  E & M & H & X & Avg.\\
    \midrule

    VAD & \cite{jiang2023vad} & 
    51.3 & 31.1 & 25.3 & 26.5 & 31.4 &
    36.3 & \phantom{0}9.5 & \phantom{0}8.0 & 11.5 & 13.4 \\

    LTF & \cite{chitta2022transfuser} &
    67.8 & 35.1 & 26.2 & 40.5 & 38.9 &
    58.9 & 18.0 & \phantom{0}9.8 & 25.9 & 23.7 \\

    UniAD & \cite{hu2023uniad} &
    78.4 & 60.5 & 33.6 & 17.8 & 45.9 &
    64.9 & 45.8 & 20.6 & \phantom{0}6.6 & 32.7 \\

    
    \rowcolor{blue!15}
    \method{} &  & 80.9 & 50.5 & 33.8 & 47.1 & \textbf{49.8} 
    & 73.3 & 34.6 & 18.8 & 32.5 & \textbf{35.7} \\

    \bottomrule
    \multicolumn{12}{r}{\textit{\small E: Easy, M: Medium, H: Hard, X: Extreme}}
    \end{tabular}
    \caption{\textbf{Photorealistic closed-loop evaluation on HUGSIM \citep{zhou2024hugsim}}. Zero-shot generalization using the \method model from the NAVSIM-v1 evaluation. Scores are per difficulty and overall average road completion (RC) and HD-Score, higher always better.}
    \label{tab:hugsim}
\end{table}

%% file: tables/navsim_v2_small.tex
\begin{table*}[t]
    \centering
    \small
    \setlength{\tabcolsep}{1.8pt}
    \begin{tabular}{@{}l@{\hspace{0.1cm}}r|ccccccccc|ccccccccc|c@{}}
    \toprule
            & & \multicolumn{9}{c}{Stage 1} & \multicolumn{9}{c}{Stage 2} & \\
    Method & &NC & DAC & DDC & TLC & EP & TTC & LK & HC & EC & NC & DAC & DDC & TLC & EP & TTC & LK & HC & EC & EPDMS \\
    \midrule
    RAP-DINO \tiny{(ViT-H)} & \cite{feng2025rap} & 97.1 &94.4 &98.8& 99.8& 83.9& 96.9 &94.7 &96.4 &66.2& 83.2&83.9& 87.4 &98.0 &86.9 &80.4 &52.3 &95.2& 52.4 &39.6\\
    GTRS-D \tiny{(V2-99)} & \cite{li2025generalized} &
     98.9  &96.2  &99.4 & 99.3 & 72.9 & 98.9 &95.1  &96.9 & 39.1 &
    91.2 & 89.4 &94.4 & 98.8 &69.5  &  90.0 & 54.3 &94.0  & 48.7& 45.0\\
    GTRS-A \tiny{(V2-99)} & \cite{li2025generalized} &
      98.9  &95.1& 99.1& 99.6 & 76.2 & 99.1  & 94.9 &97.6& 54.2 &88.1& 
      88.8 &89.3 &  98.9& 98.9&  85.9& 53.7&  96.8&56.9& 45.4\\

    \rowcolor{blue!5}
    GTRS-DrivoR \tiny{(ViT-S)}$^{*}$ &  &  98.0    &  95.8   &   99.7      &99.3       &72.9    &98.2     & 95.6     & 96.9   &   51.6    & 91.6    &  86.7  & 90.2   &98.8 &73.2     & 88.9     &51.9    & 94.9       &46.4 &45.8\\

    ZTRS \tiny{(V2-99)} & \cite{li2025ztrs} & 98.9 & 97.6 &100& 100 &66.7& 98.9& 96.2& 96.7& 44.0 & 91.1 & 90.4& 95.8& 99.0 &63.6& 89.8 &60.4& 97.6& 66.1& 48.1 \\
    \rowcolor{blue!15}
    \method{} \tiny{(ViT-S)} & &98.8& 95.1& 98.9& 100& 72.6&98.7&94.0&97.6&73.3&90.2&88.4&91.9&98.6&70.0&88.0&50.1&98.5&76.2&  48.3\\

    \rowcolor{blue!15}
    \method{} \tiny{(+65k SimScale data, ViT-S)} & &98.9& 97.3& 99.2& 99.6& 77.7&99.1&95.3&97.6&68.4&92.3&92.2&97.0&99.0&72.1&90.3&56.3&97.1&38.8&  52.3\\

    SimScale \tiny{(+185k SimScale data, V2-99)}& \cite{tian2025simscale} & 99.6	& 99.16	& 99.9	& 100	& 69.6	& 99.6	& 95.8	& 95.6	& 28.4 & 94.5	&94.2	&95.8	&99.2&	75.8	&92.8	&60.1	&96.1	&43.2 & 53.2 \\

    \rowcolor{blue!15}
    \method{} \tiny{(+134k SimScale data, ViT-S)} & &99.1& 98.2& 99.3& 99.8& 75.4&98.7&94.9&97.6&70.2&92.3&91.6&97.3&99.1&75.7&90.6&56.1&98.4&44.7& \bf 54.6\\

    \bottomrule
    \multicolumn{21}{@{}l@{}}{$^*$: \scriptsize{uses the same ViT-S backbone with registers as \method, the prediction and scoring heads remain the same as in GTRS.}}
    \end{tabular}
    \vspace{-7pt}
    \caption{\textbf{NAVSIM-v2 \texttt{navhard-two-stage}}. Comparison to existing methods on the NAVSIM-v2 benchmark test set using the EPDMS. Full definition of all scores in supplementary material, larger always better.
    Note: other scores reported in the literature before the official benchmark bug fix of the metrics are reported in supplementary material.
    }
    \label{tab:benchmark_navsim_v2_small}
\end{table*}

%% file: tables/ablations_perception_all.tex
\begin{table}[ht]

    \begin{subtable}{\linewidth}
    \centering
    \small
    \begin{tabular}{c|ccb}
    \toprule
    Init. & Random & ImageNet 21k & DINOv2 \\
    \midrule
    PDMS  & 70.1 & 87.5 & 90.0 \\
    \bottomrule
    \end{tabular}
    \caption{\textbf{Pretaining.}}
    \label{tab:ablations_pretraining}
    \end{subtable}
    
    \vspace{4pt}
    \begin{subtable}{\linewidth}
    \centering
    \small
    
    \setlength{\tabcolsep}{3pt}
    \begin{tabular}{@{}llll|l|cc|c@{}}
    \toprule
    & Compres- & Cam. & Scene & Img. Enc. & \multicolumn{2}{c|}{Parameters} & \\
    & sion     & tokens & tokens & training   & Optim & Total & \multirow{-2}{*}{PDMS}\\
    \cmidrule{2-8}
    \rowcolor{black!10}
    (a) &  & & & Frozen & 18.2 & 41.2 & 88.2 \\ 
    \rowcolor{black!10}
    (b) & \multirow{-2}{*}{No} & \multirow{-2}{*}{4k} & \multirow{-2}{*}{16k} & LoRA   & 18.8 & 41.8 & 90.2 \\
    \cmidrule{2-8}
    (c) & Pooling & 16 & 64 & LoRA & 18.2 & 41.2 & 89.7  \\
    \cmidrule{2-8}
    (d) & & 16 & 64 & Frozen & 19.3 & 42.3 &  86.9 \\ 
    (e) & \multirow{-2}{*}{Decoder} & 16 & 64 & LoRA & 19.9 & 42.9 & 89.3  \\
    \cmidrule{2-8}
    (f) & & 16 & 64 & Full ft.  & 41.2 & 41.2 & 88.4 \\
    (g) & & 16 & 64 & Frozen & 18.2 & 41.2 &  84.4 \\
    \rowcolor{blue!15}
    (h) & \multirow{-3}{*}{Registers} & 16 & 64 & LoRA        & 18.8 & 41.8 & \bf 90.0 \\
    \bottomrule
\end{tabular}

\caption{\textbf{Compression and finetuning.} Note the higher performance of register based compression, nearing the model using 250x as many tokens.}
\label{tab:ablations_perception_backbone}
\end{subtable}
    
    \vspace{4pt}
    \begin{subtable}{\linewidth}
    \centering
    \small
    \setlength{\tabcolsep}{3pt}
    \begin{tabular}{cc|l|c}
    \toprule
    \multicolumn{2}{c|}{\# Tokens} & \multirow{2}{*}{Registers} & \multirow{2}{*}{PDMS} \\
    per cam. & per scene & & \\
    \midrule
    5 & 20 & DINOv2 & 88.1 \\
    5 + 16 & 84 & DINOv2 + Rand. init. & 89.8 \\
    \midrule
     5 & 20 & Rand. init. & 89.7  \\
     8 & 32 & Rand. init. & 89.7  \\
    \rowcolor{blue!15}
     16 & 64&  Rand. init. & \bf 90.0 \\
     32 & 128&  Rand. init. & 89.8 \\
    \bottomrule
    \end{tabular}
    \caption{\textbf{Scene token} count influence. }
    \label{tab:ablations_num_tokens}
\end{subtable}

\caption{\textbf{Ablations of the perception.} All results are presented on \texttt{navval} using a ViT-S backbone.}

\end{table}

%% file: figures/token_qualitative/token_qualitative.tex
\begin{figure}[t]
    \centering
    \small
    \includegraphics[width=\linewidth]{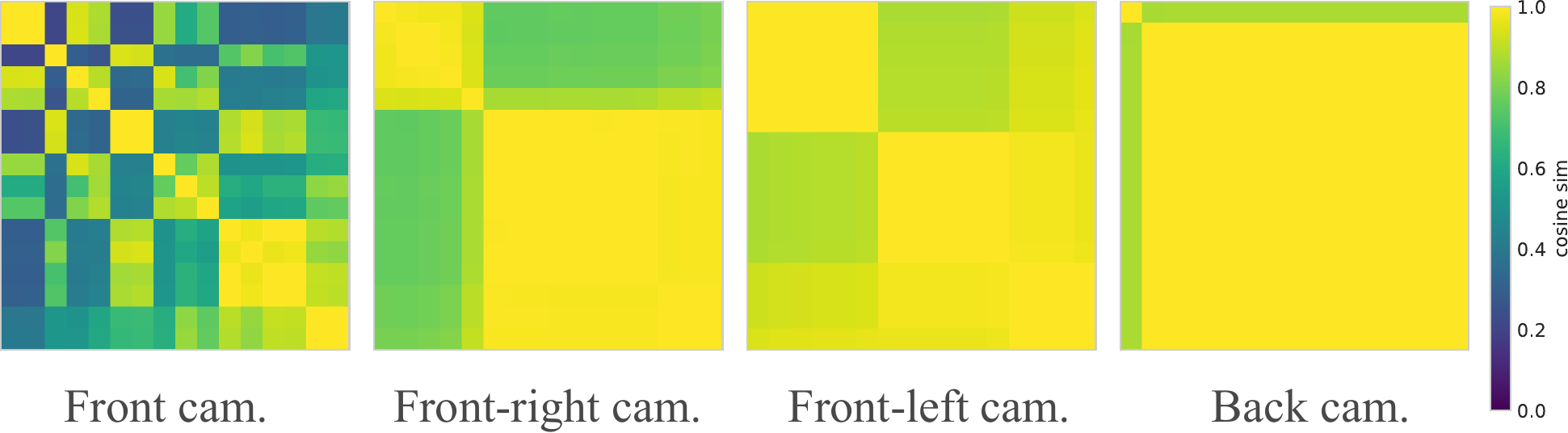}
    \caption{\textbf{Cosine similarity between scene tokens.} Darker indicates lower cosine similarity. Note the specialized tokens in the front cam, and collapsed tokens in the back cam, showing relative camera compression. Averaged on \texttt{navval}.}
    \label{fig:token_qualitative}
\end{figure}

%% file: figures/token_qualitative/token_attention_maps.tex
\begin{figure*}[t]
    \centering
    \setlength{\tabcolsep}{1pt}
    \begin{tabular}{cc|cccc}
        \rotatebox{90}{\quad Front cam.} &
        \includegraphics[width=0.19\linewidth]{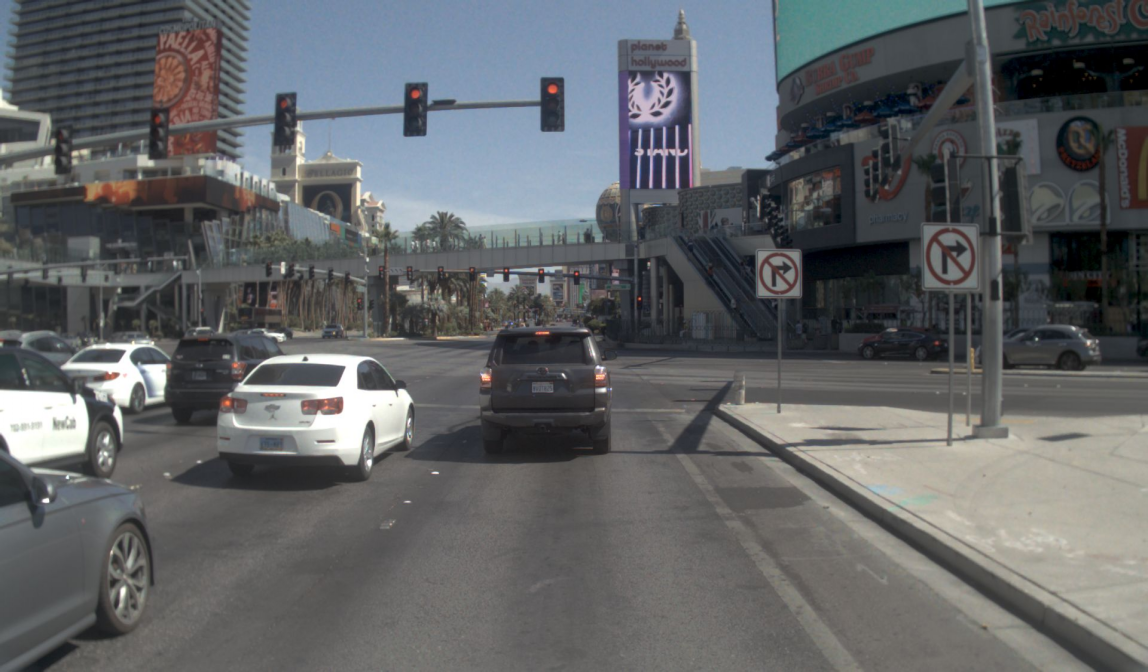} &
        \includegraphics[width=0.19\linewidth]{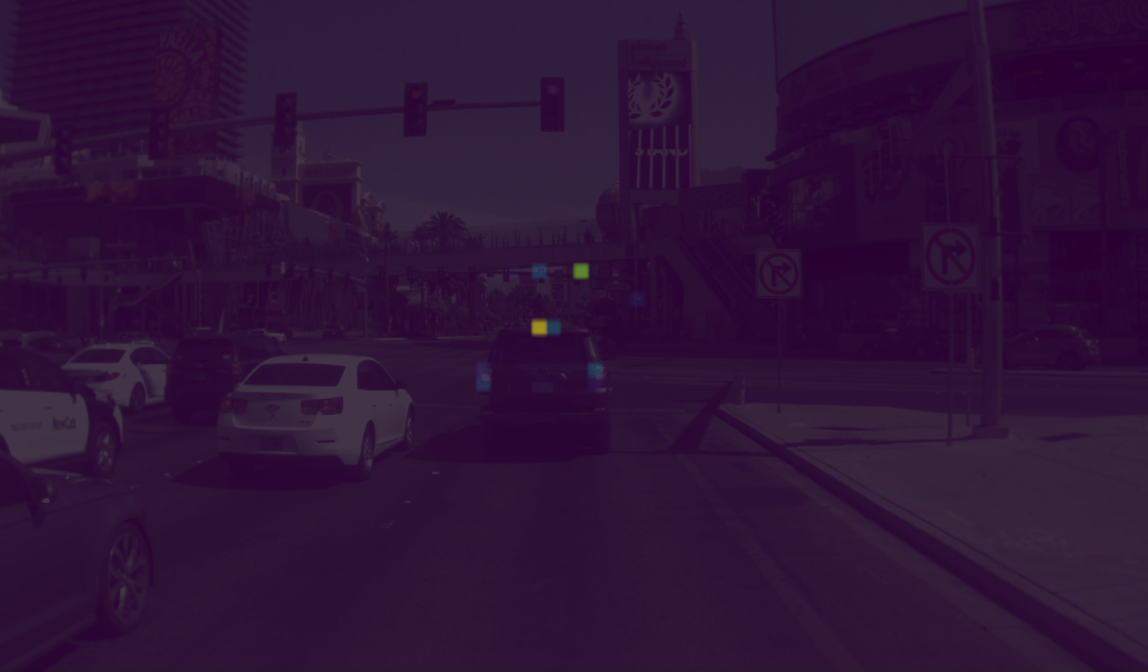} &
        \includegraphics[width=0.19\linewidth]{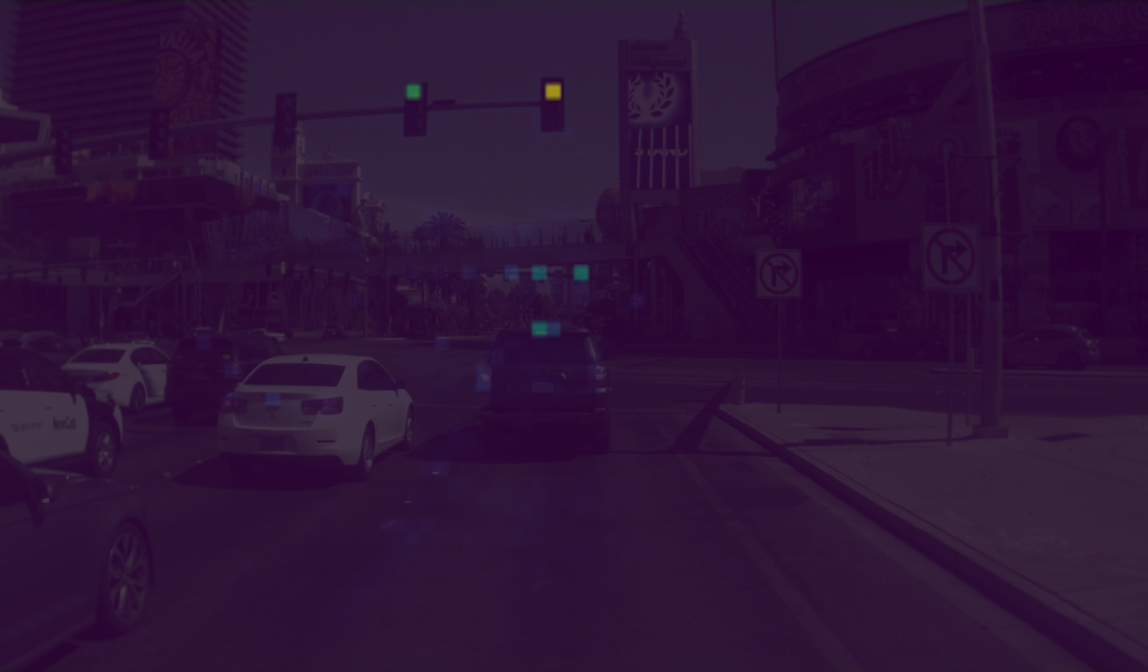} &
        \includegraphics[width=0.19\linewidth]{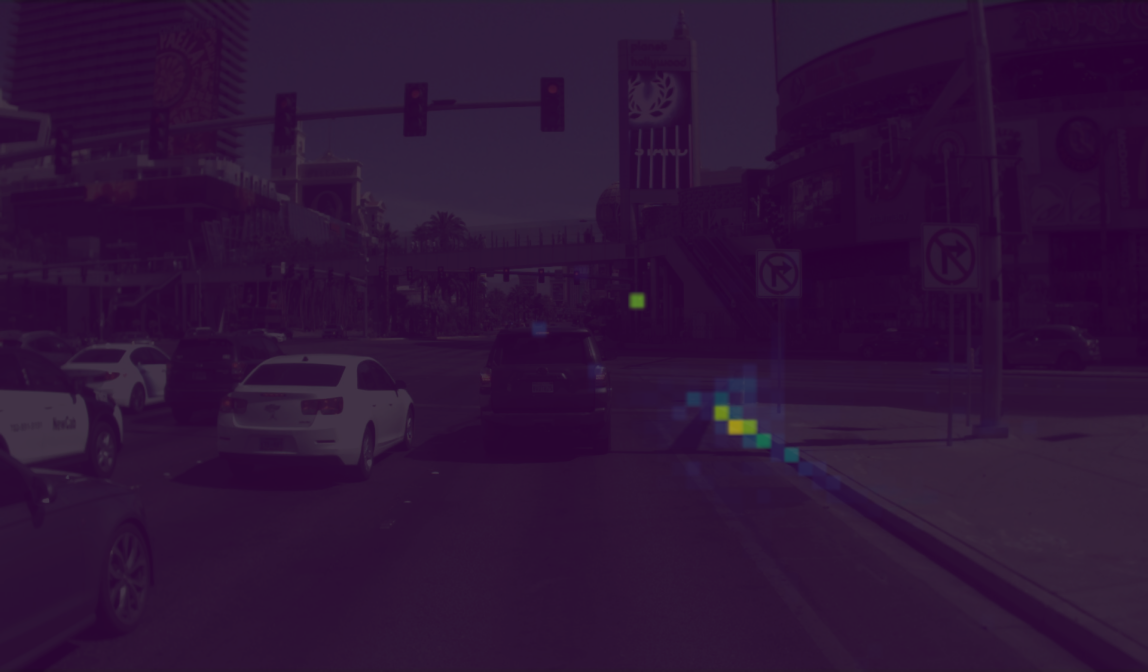} &
        \includegraphics[width=0.19\linewidth]{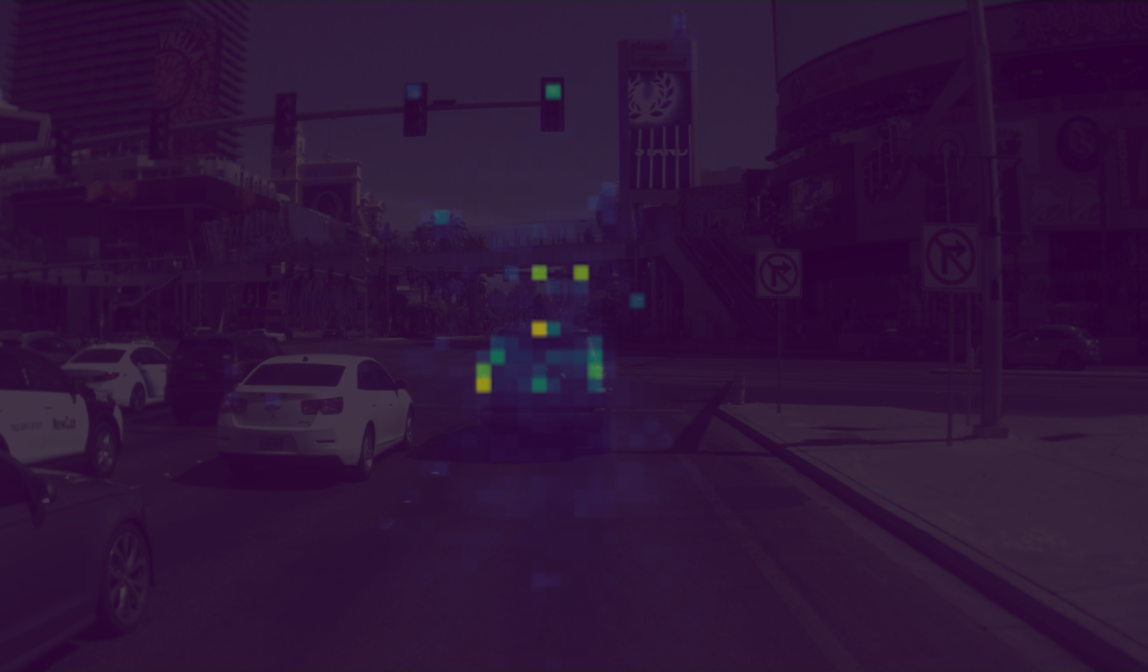} \\
        \rotatebox{90}{\quad Back cam.} & 
        \includegraphics[width=0.19\linewidth]{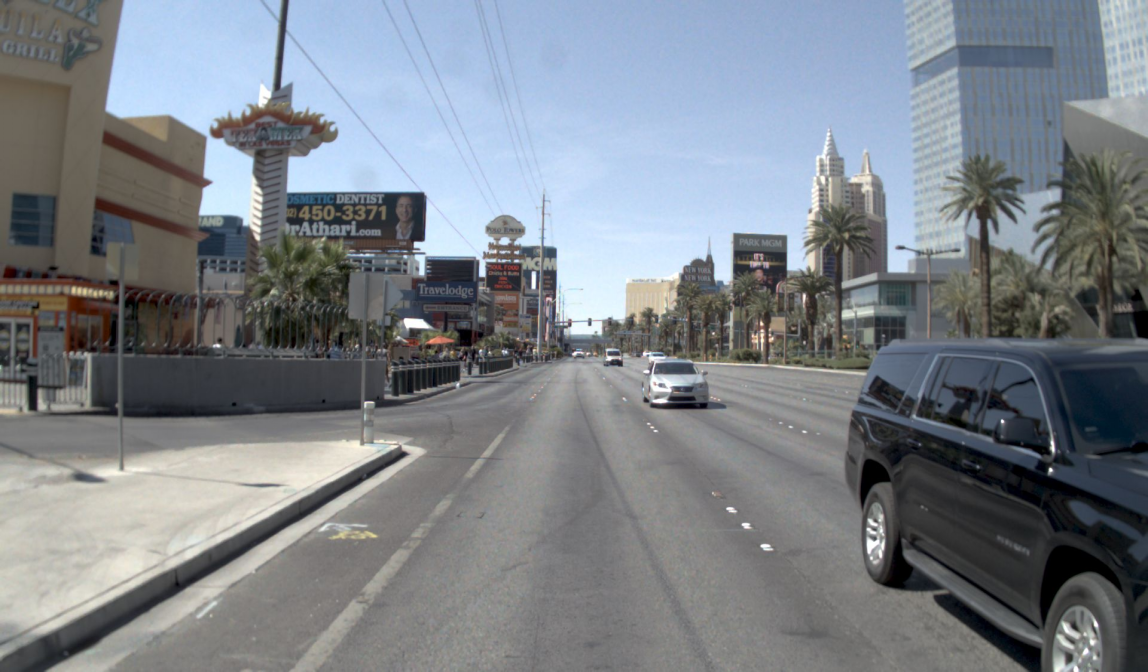} &
        \includegraphics[width=0.19\linewidth]{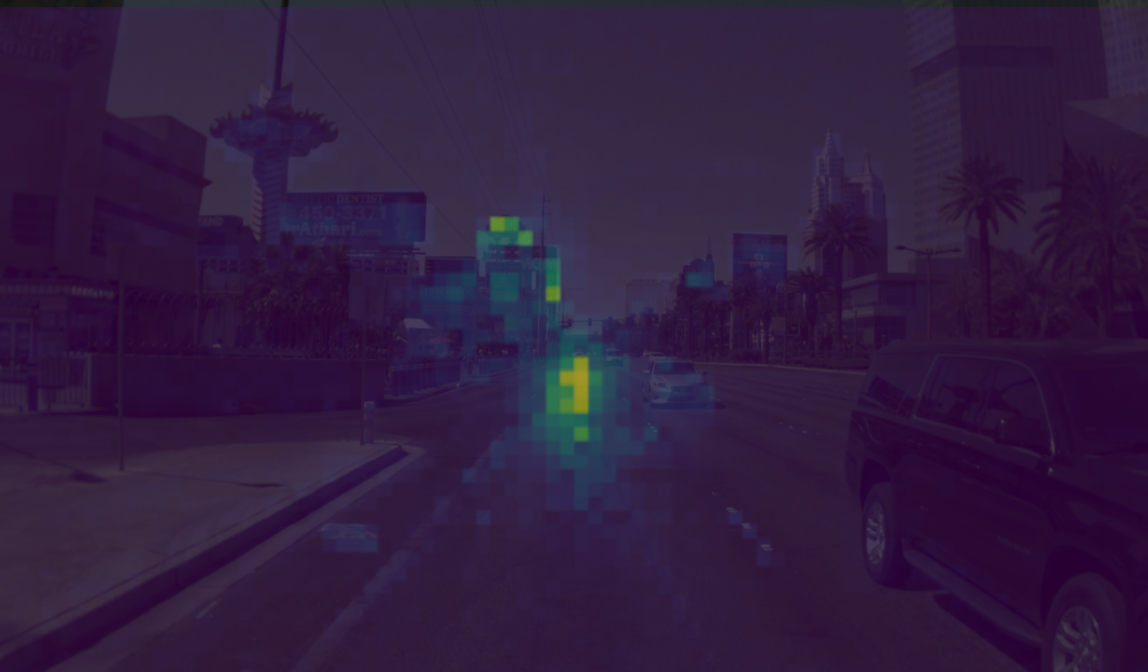} &
        \includegraphics[width=0.19\linewidth]{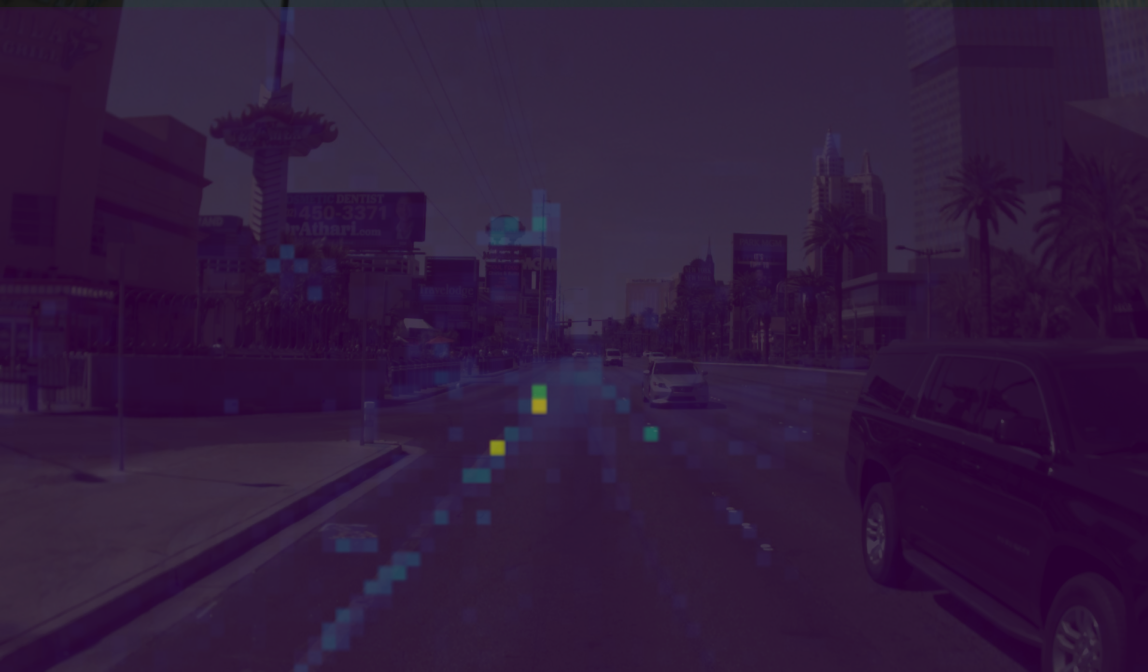} &
        \includegraphics[width=0.19\linewidth]{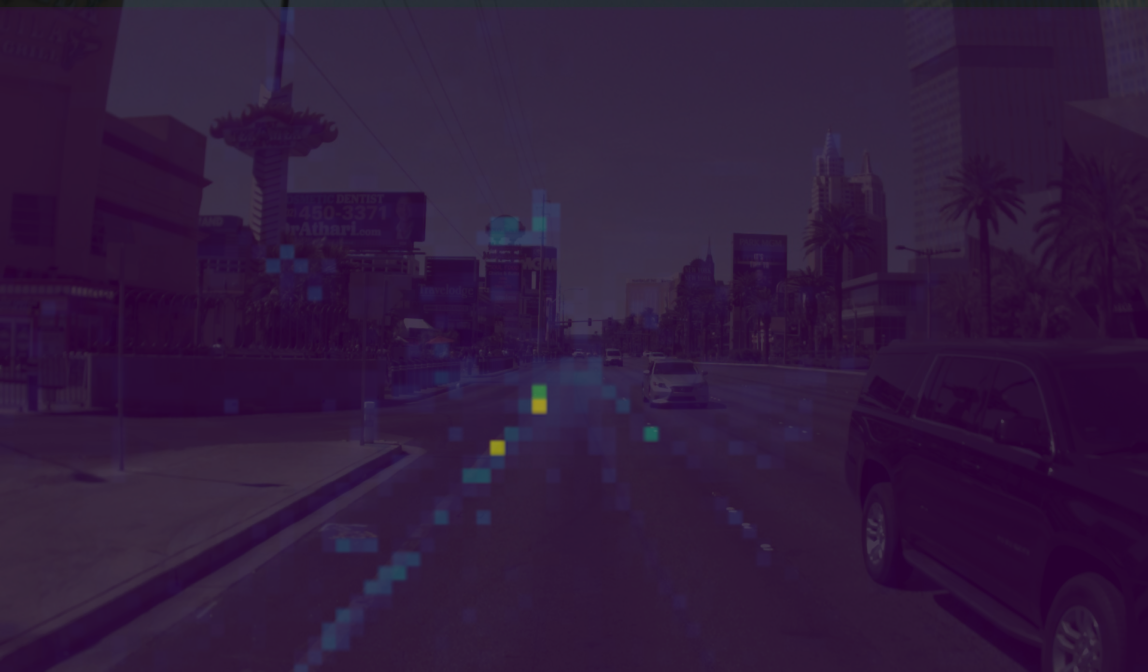} &
        \includegraphics[width=0.19\linewidth]{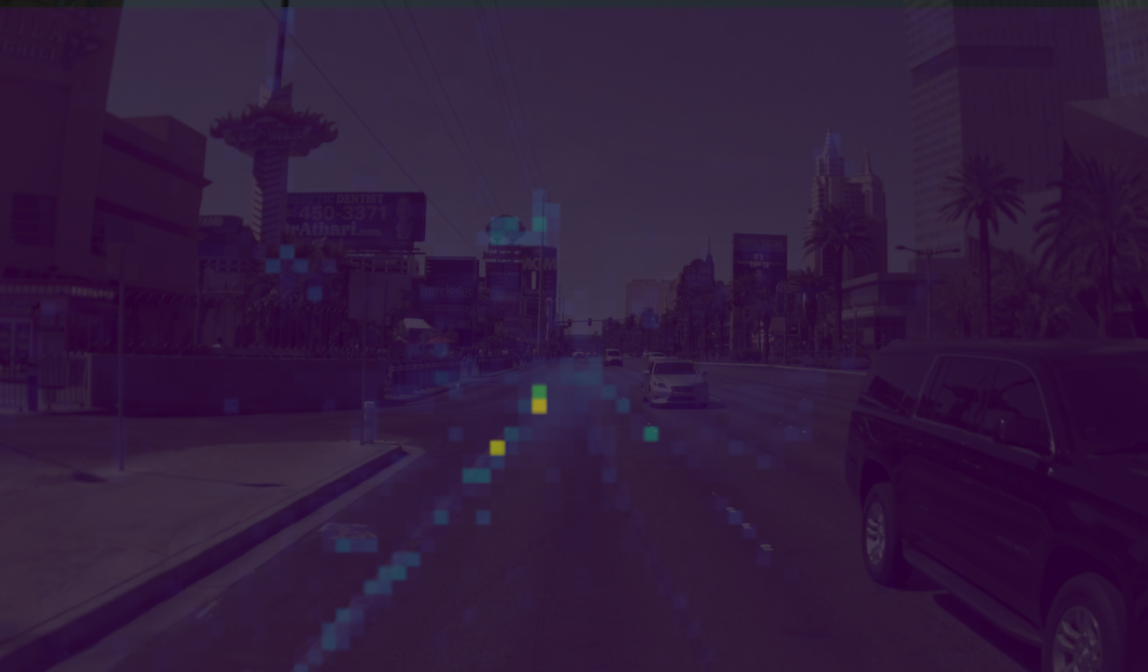} \\
    \end{tabular}
    \caption{\textbf{Attention maps of scene tokens.} From the final attention layer, front-camera tokens specialize to distinct regions (traffic light, lead vehicle, road edges), while back-camera tokens largely collapse to the same features, aside from a single distinct token.}
    \label{fig:token_attention}
\end{figure*}

%% file: tables/ablations_trajectory.tex
\begin{table}[t]
        \centering
        \small
        \begin{tabular}{c|ccccbc}
            \toprule
            Num traj. & 1 & 8 & 16 & 32 & 64 & 128 \\
            \midrule
            PDMS      & 80.1 & 87.6 & 88.1 & 89.5 & \bf 90.0 & \bf 90.0 \\
            \bottomrule
        \end{tabular}
        \caption{\textbf{Ablations of the trajectory prediction.} Influence of the number of trajectories on \texttt{navval}.}
        \label{tab:ablations_scaling_trajectories}
\end{table}

%% file: tables/ablations_scoring_tiny.tex
\begin{table}[t]
    \centering
    \small
    \setlength{\tabcolsep}{2pt}
    \begin{tabular}{@{}lc|c|c||cc@{}}
    \toprule
    & Separate &  Disentan-      & Score & \multirow{2}{*}{PDMS} & Behavior\\
    & decoders &  glement  & num.  &      & control\\
    \midrule
    (a) & \xmark   & - & 6 & 84.7 & \cmark \\
    (b) & \cmark   & \xmark  & 6 & 86.8 & \cmark \\
    \rowcolor{blue!15}
    (c) & \cmark   & \cmark & 6 & \bf 90.0 & \cmark\\
    \rowcolor{black!10}
    (d) & \cmark   & \cmark & 1 & 88.2 & \xmark\\
    \bottomrule
    \end{tabular}
    \caption{\textbf{Ablations of the scoring} on \texttt{navval}. \emph{Disentanglement} refers to a stop-gradient followed by 
    an embedding of decoded trajectories. \emph{Behavior control} stands for predicting all 6 PDMS components.}
    \label{tab:ablations_scoring}
\end{table}

%% file: figures/scoring_qualitative/scoring_qualitative.tex
\begin{figure}
    \centering
    \includegraphics[width=\linewidth]{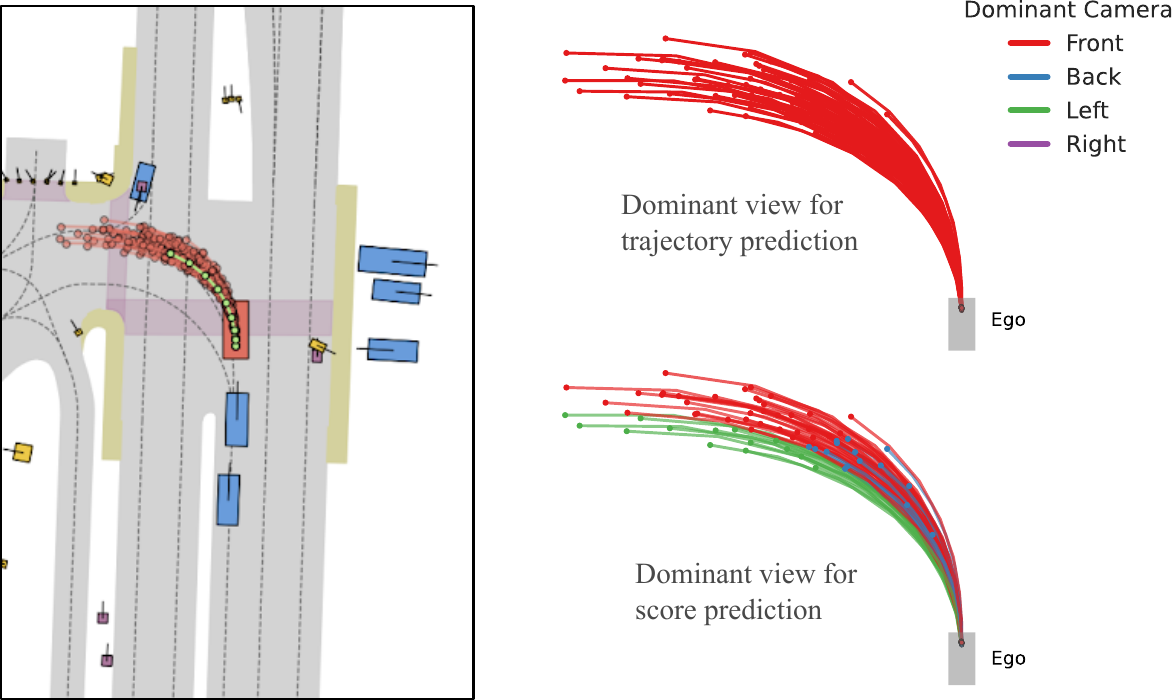}
    \caption{\textbf{Scoring head disentanglement.} Dominant cameras are identified via cross-attention between scene tokens and score or trajectory tokens. Trajectory prediction consistently relies on the front camera, while scoring shifts attention based on trajectory behavior—underscoring the need to separate the two pipelines.}
    \label{fig:scoring_qualitative}
\end{figure}

%% file: tables/ablations_loss_long_traj.tex

\begin{table}[t]
        \small
        \centering
        \setlength{\tabcolsep}{3pt}
        \begin{tabular}{@{}l|cc@{}}
            \toprule
            Targets &  \texttt{navval} (PDMS) & \texttt{warmup} (EPDMS)\\
            \midrule
            ($t+T$) & 90.0 &  \bf 39.4 \\
            ($t+T$) \& ($t+T'$) & \bf 90.6 & 37.8 \\
            \bottomrule
        \end{tabular}
        \caption{\textbf{Ablations.} Multiple targets for WTA regression, we report PDMS score for NAVSIM-v1 \texttt{navval} and EPDMS for NAVSIM-v2 \texttt{warmup-two-stage}. \label{tab:ablations_loss_long}}
\end{table}

%% file: figures/safety_agent/safety_agent.tex
\begin{figure}
    \centering
    \includegraphics[width=\linewidth]{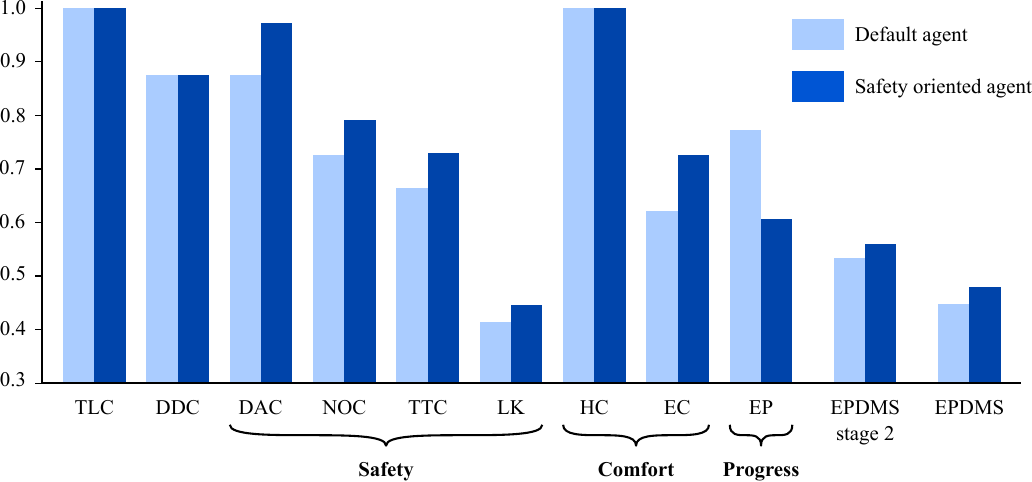}
    \caption{\textbf{Safety-oriented agent.} Dark blue was tuned on \texttt{warmup-two-stage}, light blue is our NAVSIM-v1 model. The result of our behavior tuning is a more cautious, but safer agent.}
    \label{fig:safety_agent}
\end{figure}

%% file: sec/5_conclusion.tex
\section{Conclusion}
\label{sec:conclusion}
We present \method, a novel E2E driving method using register-based compression and disentangled scoring representations. 
\method highlights that full-transformer architectures without complex intermediate states nor large trajectory dictionaries can achieve state-of-the-art results. 
Future works may explore compression incorporating historical frames, additional sensors or map information.


%% file: sec/X_suppl.tex
\clearpage
\setcounter{page}{1}
\maketitlesupplementary
\setcounter{table}{7}
\setcounter{figure}{6}
\setcounter{equation}{4}

\appendix

\section{Description of PDMS and EPDMS metrics}

The \emph{Predictive Driver Model Score (PDMS)}, main metric of NAVSIM-v1, and the \emph{Extended  Predictive Driver Model Score (EPDMS)}, main metric of NAVSIM-v2, follow a similar definition, based on a group $G$ of sub metrics.
$G$ can be divided into two kinds of metrics: a group $G_p$ of penalties, and a group $G_b$ of behavioral metrics. 
$G_p$ contains the metrics measuring the compliance to driving rules, such as drivable area compliance or collision occurrence.
$G_b$ contains the metrics measuring more driver-related metrics, such as comfort or progress towards the goal.
The PDMS and EPDMS metrics are defined as:
\begin{equation}
    \text{Score}(\tau_i) = \prod_{c \in G_p} \mathcal{G}_c(\tau_i)^{\kappa^p_c} \times \frac{1}{Z} \sum_{c \in G_b}{\kappa^b_c \; \mathcal{G}_c(\tau_i)}
\end{equation}
for $\tau_i$ a candidate trajectory, $Z=\sum_{c \in G_b}{\kappa^b_c}$ and $\kappa$ as defined in \autoref{tab:full_hparams}.
$G_p$ metrics are multiplicative, meaning one failure to comply to a targeted driving rule, i.e., $\mathcal{G}_c(\tau_i)=0$ results in a score equal to zero.
On the contrary, $G_b$ metrics are additive and allow compromise between the metrics, e.g., comfort and progress.

\begin{table}[ht]
\centering
\small
\setlength{\tabcolsep}{3pt}
\begin{tabular}{@{}lllcc@{}}
\toprule
Abbr. & Sub-score & Grp. & PDMS & EPDMS\\
      & $c$                     &      & $\kappa_c$ & $\kappa_c$ \\
\midrule
NC   & No-at-fault Collisions  & $G_p$ & 1 & 1\\
DAC  & Drivable Area Compliance & $G_p$ & 1 & 1\\
DDC  & Driving direction Compliance & $G_p$ & 0 & 1\\
TLC & Traffic-line compliance & $G_p$ & - & 1 \\
TTC  & Time to Collision & $G_b$ & 5 & 5\\
EP   & Ego Progress & $G_b$ & 5 & 5\\
Comf.   & Ego Comfort & $G_b$ & 2 & - \\
LK   & Lane Keeping & $G_b$ & - &2\\
HC   & History Comfort & $G_b$ & - &2\\
EC   & Extended Comfort & $G_b$ & - &2\\
\bottomrule
\end{tabular}
\caption{\textbf{Weights of the sub-scores} in the PDMS and EPDMS.}
\label{tab:full_hparams}
\end{table}

\section{Training loss weights}

We recall here the equation of the score loss defined in the main paper: 
\begin{equation}
    \mathcal{L}_\text{score} =  \sum_{c}  \lambda_c  \sum_i \operatorname{BCE}\left(\mathcal{G}_{\theta_c}(\tau_i), \mathcal{G}_c(\tau_i) \right)
\end{equation}
which corresponds an individual loss for each sub-score of the PDMS.
During training, the predicted sub-scores and their associated weight $\lambda_c$ are set as defined in \autoref{tab:placeholder}, i.e., all $\lambda_c$ are set to~1.

\begin{table}[]
    \centering
    \small
    \setlength{\tabcolsep}{3pt}
    \begin{tabular}{c|cccccc}
    \toprule
    Sub-score & NC &  DAC & DDC & TTC & EP & Comf.\\
    \midrule
    $\lambda_c$ & 1 & 1 & 1 & 1 & 1 & 1 \\
    \bottomrule
    \end{tabular}
    \caption{\textbf{Training loss weights of the sub-scores.}}
    \label{tab:placeholder}
\end{table}

Similarly, as defined in the main paper, the final loss is:
\begin{equation}
    \mathcal{L} = \mathcal{L}_\text{traj} + \lambda_s \mathcal{L}_\text{score}.
\end{equation}
where, again, $\lambda_s$ is set to 1.

\section{Inference weights of the sub-scores}

\begin{table}[ht]
\centering
\small
\setlength{\tabcolsep}{3pt}
\begin{tabular}{@{}lllccc@{}}
\toprule
Abbr. & Sub-score & Grp. & PDMS & EPDMS\\
               & $c$                    & & $\kappa_c$  & $\kappa_c$\\
\midrule
NC   & No-at-fault Collisions  & $G_p$ &1 & 10\\
DAC  & Drivable Area Compliance & $G_p$ & 1 & 13\\
DDC  & Driving direction Compliance & $G_p$ & 0 & 6\\
TTC  & Time to Collision & $G_a$ & 5 & 14\\
EP   & Ego Progress & $G_a$ & 5 & 15\\
Comf.   & Ego Comfort & $G_a$ & 2 & 2\\
\bottomrule
\end{tabular}
\caption{\textbf{Inference weights of the sub-scores.}}
\label{tab:score_def}
\end{table}

The subscore prediction design in \method enables flexible weighting strategy. In~\autoref{tab:score_def}, we provide the sub-score weights for both PDMS (NAVSIM-v1) and EPDMS (NAVSIM-v2). For simplicity, the model trained for NAVSIM-v1 keeps the same weights as in the standard PDMS metrics~\cite{dauner2024navsim}. For NAVSIM-v2, we adjust the weights with the exception of the Comfort. The adjustment is validated with NAVSIM-v2 \texttt{warmup-two-stage}, see Figure 6 in the original paper. 

\input{tables/efficiency_rebuttal}

\section{Implementation details}
We conduct our experiments on a node of 8\,$\times$\,A100 GPUs. We run each training on 4\,$\times$\,A100 GPUs with a base learning rate of 0.0002 and batch size of 16, using AdamW optimizer and a cosine annealing learning rate scheduler. The training lasts around 1.5 hours per epoch for the competition split (\texttt{navtrain} + \texttt{navval}) and around 1 hour per epoch for \texttt{navtrain} only. As described in the main paper, we train for 25 epochs for the NAVSIM-v1 (\texttt{navtrain} + \texttt{navval}) and 10 epochs for NAVSIM-v2 \texttt{navtrain}.

\section{Efficiency analysis}~\label{supp_sec:efficiency}
\autoref{tab:efficiency} shows the full results of the throughput analysis. We benchmarked each model on a single batch on a single A100 GPU. FLOPs counts were conducted using the FVCore library. Peak memory was counted using Pytorch's \texttt{max\_memory\_allocated} function. (Note: FVCore currently excludes Scaled Dot Product Attention and thus FLOPs counts do not reflect the real count for each model, though relative performance should remain consistent.) Throughput was counted on a single forward pass after three warm-up iterations, with the final number averaged over 10 iterations. \method has much improved throughput and lower memory consumption, demonstrating the best tradeoff between performance and efficiency.

\section{Single- vs multi- token per trajectory}

\begin{figure}[ht!]
    \centering
    \begin{minipage}{0.45\linewidth}
        \centering
        \includegraphics[width=\linewidth]{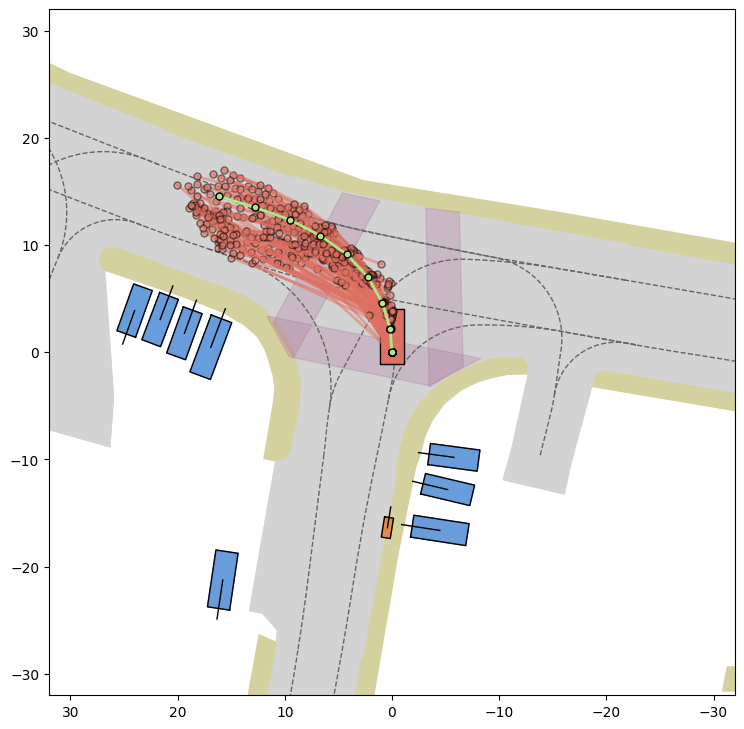}
        \subcaption{Multi-token}
    \end{minipage}
    \hfill
    \begin{minipage}{0.45\linewidth}
        \centering
        \includegraphics[width=\linewidth]{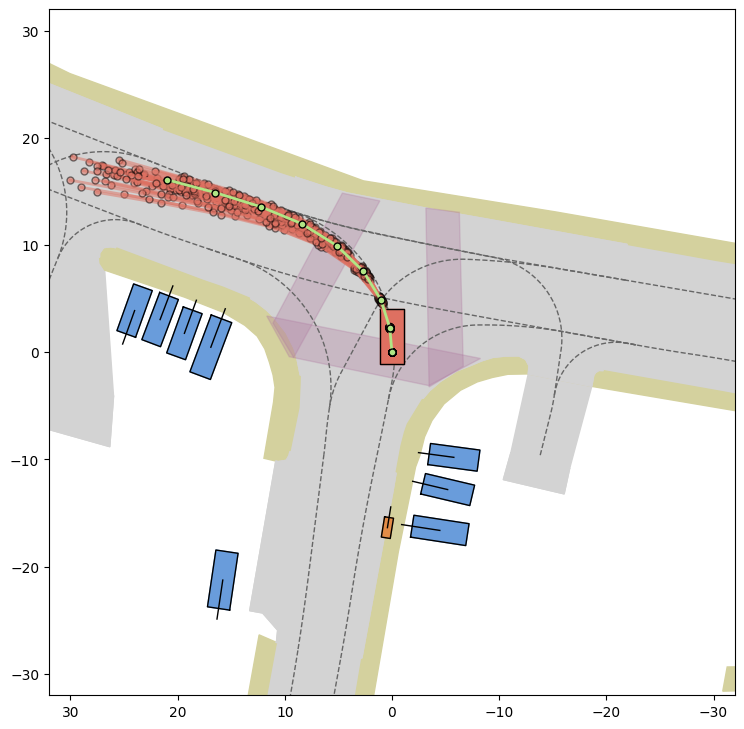}
        \subcaption{Single-token}
    \end{minipage}
    \caption{\textbf{Single- vs multi- token trajectory.} Qualitative comparison: single-token trajectories are much smoother and less noisy.}
    \label{fig:viz_single_tok}
\end{figure}

Previous works such as \cite{guo2025ipad} have used a single token per trajectory pose, decoding each token into an $(x,y,\theta)$ tuple. We ablate the choice of decoding a single token to a full trajectory vs decoding a set of tokens. \autoref{tab:ablations_trajectory_singleVSmultiple} shows that mapping a single token to a trajectory leads to a large jump in performance. \autoref{fig:viz_single_tok} visualizes the same scene, with trajectories decoded either from multi- or single-token. The trajectories mapped from a single token are much smoother and contain less noise, showing that representing trajectories with a single token simplifies learning.

\input{tables/ablations_trajectory_singleVSmultiple}

\input{tables/navsim_v1}
\section{Expanded NAVSIM-v1 results}
We provide in~\autoref{tab:benchmark_navsim_v1} a full comparison to state-of-the-art methods, including methods with LiDAR inputs and with post-processing such test-time training, Best-of-N scoring and ensembling. We note that the Best-of-N practice aims to score all predicted trajectories with ground truth and select the best trajectory based on the scores from the evaluation, which is not aligned with the evaluation protocol defined in NAVSIM~\cite{dauner2024navsim, Cao2025navsim2}.  

We note that the goal of~\method is to provide a simple and efficient baseline model with registers for end-to-end driving. Without bells and whistles, \method still achieves competitive performance compared to methods with post-processing techniques or additional sensor modality.

\section{NAVSIM-v2 results before benchmark fix}

We provide in~\autoref{tab:benchmark_navsim_v2_small_supp} the full NAVSIM-v2 state-of-the-art comparison, including the results of methods before the benchmark bug fix (Issue \#151 in NAVSIM official GitHub), which is related to the failure of filtering out human driver errors. 

From the table, we see that \method ranks among the best-performing methods, with a much lighter model design (see more detail in~\autoref{supp_sec:efficiency}) and without extra training data. 
Also, \method ranks first among state-of-the-art methods after the bug fix. This indicates that the errors made by \method are mostly due to human errors, which are not penalized after the fix according to the NAVSIM-v2 benchmark~\cite{Cao2025navsim2}.

\input{tables/navsim_v2}




\section{Limitations and additional visualizations}

\begin{figure*}[t]
    \centering
    \begin{minipage}{0.33\linewidth}
        \centering
        \includegraphics[width=\linewidth]{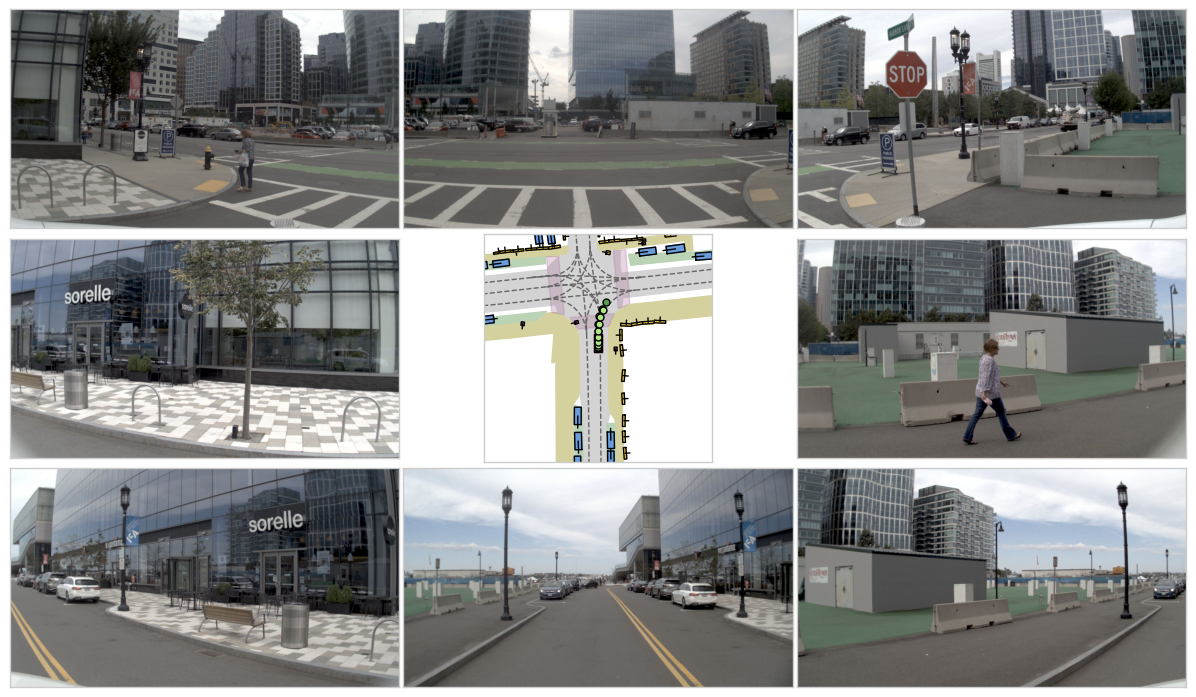}
    \end{minipage}
    \hfill
    \begin{minipage}{0.33\linewidth}
        \centering
        \includegraphics[width=\linewidth]{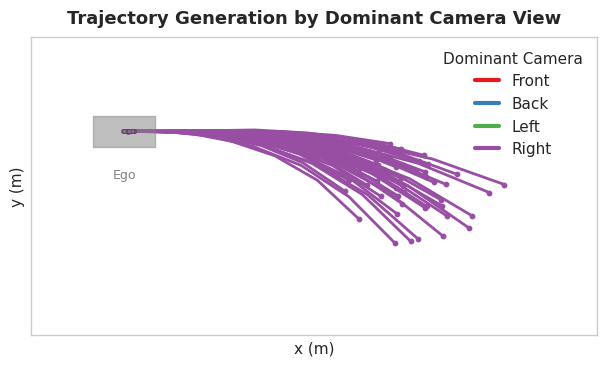}
    \end{minipage}
        \begin{minipage}{0.33\linewidth}
        \centering
        \includegraphics[width=\linewidth]{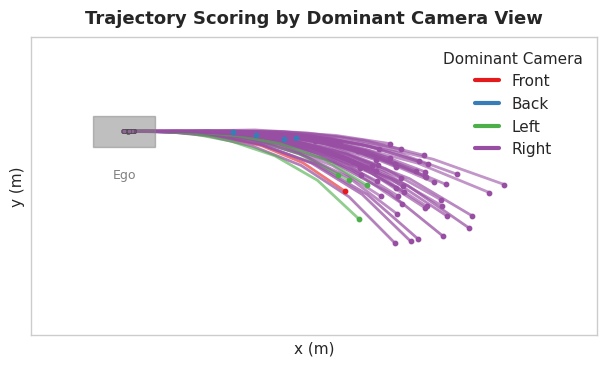}
    \end{minipage}
    \caption{\textbf{Visualization of generated trajectories (right turn) in a NAVSIM-v1 val set scene.} Right turn scenario with right camera as dominant camera used in trajectory scoring and generation.}
    \label{fig:viz_right_turn}
\end{figure*}

\begin{figure*}[t]
    \centering
    \begin{minipage}{0.33\linewidth}
        \centering
        \includegraphics[width=\linewidth]{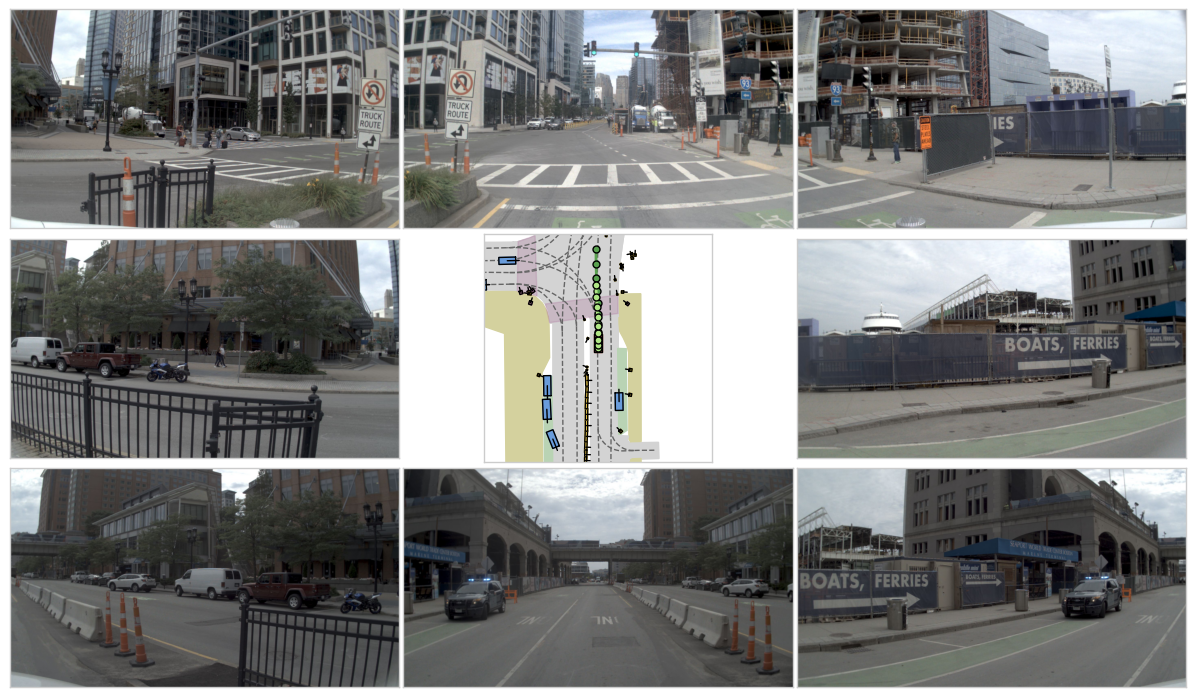}
    \end{minipage}
    \hfill
    \begin{minipage}{0.33\linewidth}
        \centering
        \includegraphics[width=\linewidth]{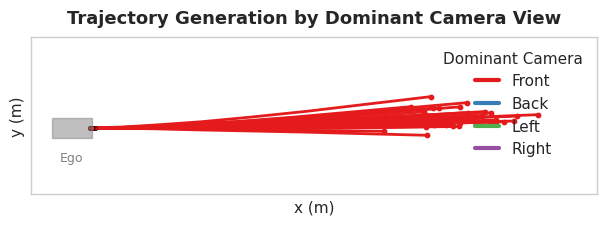}
    \end{minipage}
        \begin{minipage}{0.33\linewidth}
        \centering
        \includegraphics[width=\linewidth]{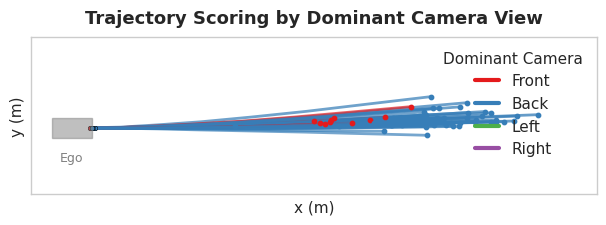}
    \end{minipage}
    \caption{\textbf{Visualization of generated trajectories (intersection crossing) in a NAVSIM-v1 val set scene.} The ego-vehicle crosses a signalized intersection, but the scoring attention focuses on the rear camera. We hypothesize that the inclusion of the Traffic Light Compliance (TLC) scoring could help in such a case. As a matter of fact, our TLC score on NAVSIM-v2 \texttt{navhard-two-stage} are high (\autoref{tab:benchmark_navsim_v2_small_supp}).}
    \label{fig:viz_no_tlc}
\end{figure*}

\begin{figure*}[t]
    \centering

    \begin{minipage}{0.49\linewidth}
        \centering
        \includegraphics[width=\linewidth]{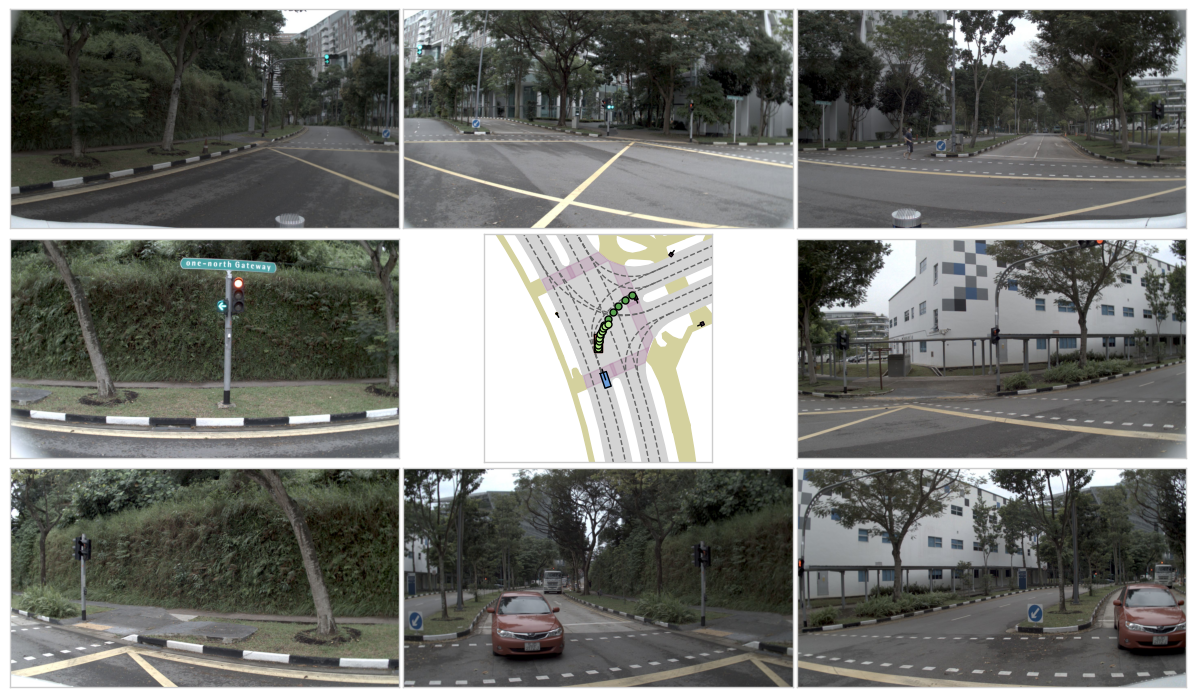}
    \end{minipage}
    \hfill
    \begin{minipage}{0.49\linewidth}
        \centering
        \includegraphics[width=\linewidth]{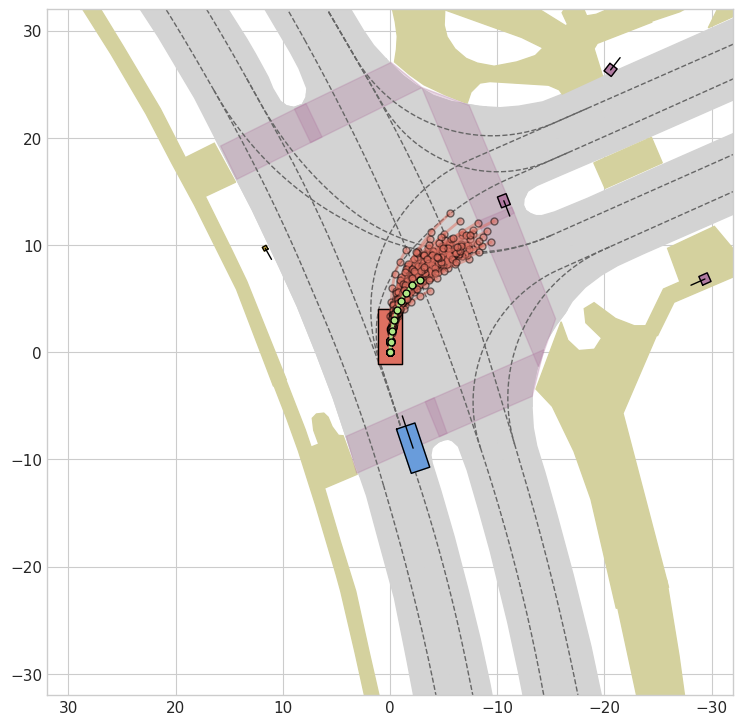}
    \end{minipage}



    \caption{\textbf{Visualization of generated trajectories (failure case) in a NAVSIM-v1 val set scene.} Failure case highlighting difficulty of navigation without historical (past) frames. The front camera image is very ambiguous, resulting in full focus on the right camera and trajectories which could result in wrong-way driving.}
    \label{fig:viz_failure_mode}
\end{figure*}

\autoref{fig:viz_right_turn}, \autoref{fig:viz_no_tlc}, and \autoref{fig:viz_failure_mode} show \method navigating diverse scenarios, and show the camera focus between the scoring and prediction pathways.

\autoref{fig:viz_no_tlc} shows a surprising case where the scoring attention is mainly focused on the back camera, despite the visible traffic light. We hypothesize that adding traffic light compliance as a score component, as in the EPDMS, could force the scoring modules attention to focus more on the traffic light as it crosses the intersection. However we highlight that despite no active scoring on this component, \method still displays very high TLC score.

\autoref{fig:viz_failure_mode} shows a failure case, highlighting the challenge of predicting viable trajectories without historical (past) frames. \method operates only on the current timestep, which makes examples like \autoref{fig:viz_failure_mode} challenging, due to the ambiguous nature of the vehicle's position in the scene. The front camera does not contain easily interpretable objects, and we see the model focusing entirely on the right camera. The resulting trajectories undercut the turn, and could result in the model driving in the wrong way. Future work could include the use of historical frames for planning, potentially alleviating these ambiguous failure cases.

\begin{figure*}[t]
    \centering
    \begin{minipage}{0.49\linewidth}
        \centering
        \includegraphics[width=\linewidth]{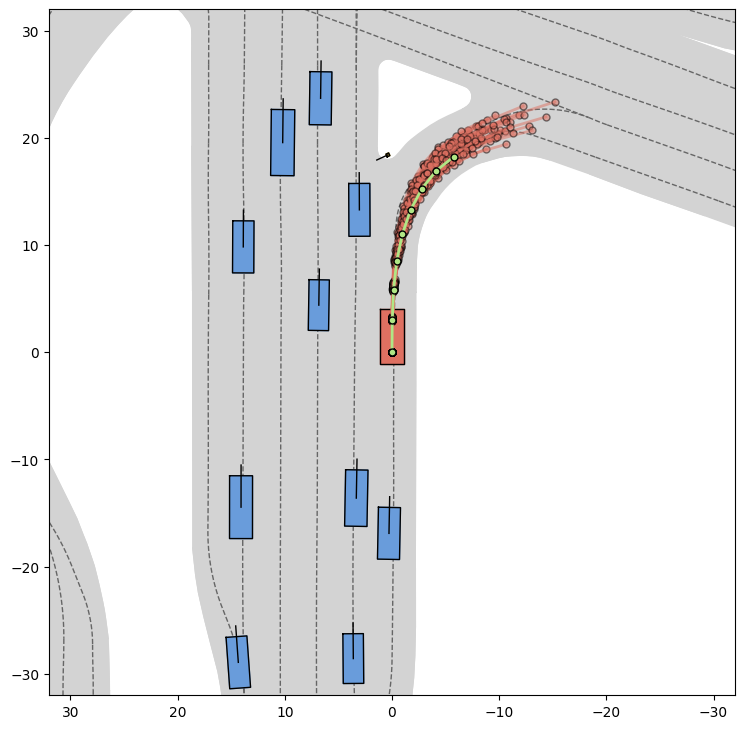}
    \end{minipage}
    \hfill
    \begin{minipage}{0.49\linewidth}
        \centering
        \includegraphics[width=\linewidth]{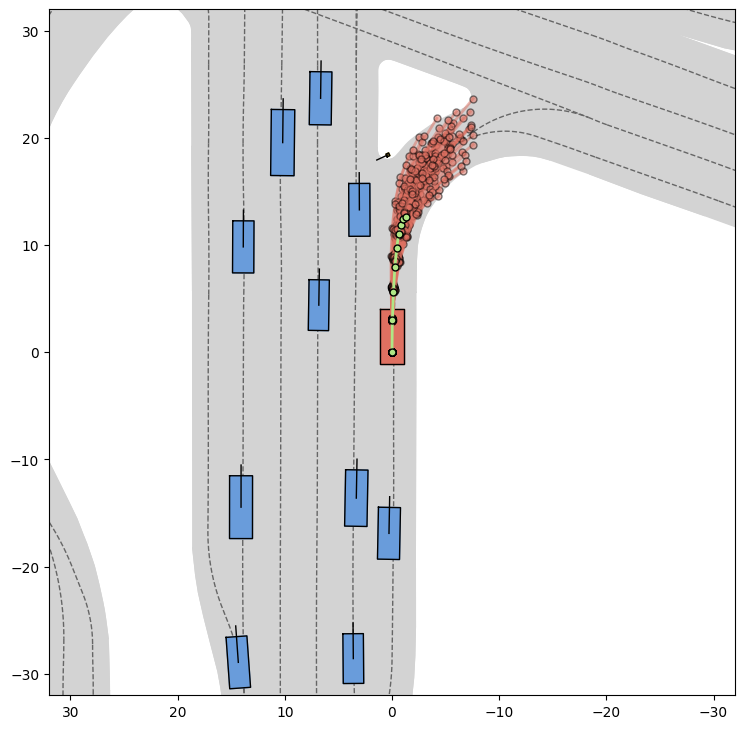}
    \end{minipage}
    \caption{\textbf{Visualization of generated trajectories (agent behavior re-weighting) in a NAVSIM-v1 val set scene.}. Left: trajectories are generated from an agent using default PDMS weights on score components. Right: the agent uses our tuned weights. Note the much shorter and less aggressive trajectories of the agent in the second (right) setting, which are better suited for navigating the out-of-distribution (OOD) scenes in the NAVSIM-v2 evaluation.}
    \label{fig:viz_agg_v_pass}
\end{figure*}

Finally, \autoref{fig:viz_agg_v_pass} compares the predicted and selected trajectories of the two models on the same scene from the NAVSIM-v1 validation set. We can see that the trajectories of our default trained NAVSIM-v1 agent are more aggressive, traveling faster and with less spread than the trajectories from our agent tuned to have more passive behavior. The ``passive'' agent is however better able to navigate the out-of-distribution (OOD) scenes in NAVSIM-v2.

%% file: tables/efficiency_rebuttal.tex
\begin{table*}[ht]
\centering
\small
\setlength{\tabcolsep}{6pt}
\begin{tabular}{ll l l|cccc|c}
\toprule
 & &   & &  &  & Peak   & Throughput & NAVSIM-v2\\
\multirow{-2}{*}{\textbf{Method}} &  \multirow{-2}{*}{Img. size} &  \multirow{-2}{*}{Cams} & \multirow{-2}{*}{Encoder} & \multirow{-2}{*}{Parameters $\downarrow$} & \multirow{-2}{*}{GFLOPs $\downarrow$} & Memory $\downarrow$ & (ms) $\downarrow$& EPDMS $\uparrow$\\
\midrule
RAP-Dino$^{\dagger}$ & (448, 768) & 4 & ViT-H & 888M & 4760 & 4.2GB & 690ms & 39.6\\ 
 &  &  &  &  \scriptsize857M$\|$31M &  \scriptsize4622$\|$138 & - &  \scriptsize653ms$\|$37ms & - \\ 
 \midrule
  
\rowcolor{black!10}
\multicolumn{9}{l}{\emph{GTRS uses extra trajectories from a diffusion-based model, introducing extra overheads.}}\\

\rowcolor{black!10}
\scriptsize +GTRS-DP   &  (512, 2048) & 4 & \scriptsize +V2-99 & \scriptsize +116M & \scriptsize +1249 & \scriptsize +1.15GB & \scriptsize +389ms & -\\
\rowcolor{black!10}
  && & \scriptsize  -& \scriptsize  110M$\|$6M & \scriptsize 770$\|$479 & \scriptsize - & \scriptsize 60ms$\|$329ms  & -\\

\midrule

\rowcolor{black!10}
GTRS-D & (512, 2048)& 1 &  V2-99 &  81M & 404& 1.0GB & 96ms & 45.0\\
\rowcolor{black!10}
- & - & - & - &   \scriptsize69.5M$\|$11.5M &  \scriptsize 345$\|$59& - &  \scriptsize 22ms$\|$74ms & -\\

\midrule

\rowcolor{black!10}
GTRS-A & (512, 2048)& 1 &  V2-99 & 171M & 439 & 1.42GB &  243ms& 45.4\\
\rowcolor{black!10}
-& -& - &  - &  \scriptsize69.5M$\|$101.5M  &  \scriptsize 345$\|$94 & - &  \scriptsize 22ms$\|$221ms& -\\
\midrule

\rowcolor{black!10}
GTRS-D$^{\ddagger}$ & (512, 2048) & 1 & ViT-L  & 321M & 1730 & 1.6GB & 400ms & 47.0\\
\rowcolor{black!10}
 & - & - & - & \scriptsize313M$\|$8M & \scriptsize 1610$\|$120 & - & \scriptsize352ms$\|$48ms & -\\

\midrule
\rowcolor{black!10}
GTRS-D$^{\ddagger}$ &(512, 2048) & 1 & ViT-S &  32M & 234& 0.38GB & 303ms & 45.9\\
 & & &   - &  \scriptsize 24M$\|$8M & \scriptsize 117$\|$117 & \scriptsize - &  \scriptsize 46ms$\|$257ms &  -\\

\midrule
ZTRS & (512, 2048) & 2 & V2-99 & 81M & 840 & 1.0GB & 193ms & 48.1\\
&  & & - &  \scriptsize69.5M$\|$11.5M  & 690$\|$150 & - & 119ms$\|$74ms & -\\
\midrule

\rowcolor{blue!15}
\method           & (672, 1148)& 4 & ViT-S   & \bf 41M & \bf 351 & \bf 0.5GB & \bf 110ms & \bf 48.3\\
     & & &   - &  \scriptsize 24M$\|$17M & \scriptsize 350$\|$1 & \scriptsize - &  \scriptsize 107ms$\|$3ms &  -\\
\bottomrule
\multicolumn{6}{@{}l@{}}{$^\dagger$: \scriptsize{RAP is trained on a dataset that is 10$\times$ larger than navtrain (the default training set).}} \qquad $^\ddagger$: \scriptsize{Reproduced.}\\
\end{tabular}
\caption{\textbf{Efficiency.} We compare the number of parameters, GFLOPs, peak memory consumption, and throughput \wrt the NAVSIM-v2 EPDMS performance. \textit{GTRS-A} refers to \textit{GTRS-Aug} and \textit{GTRS-D} for \textit{GTRS-Dense}, we decompose the parameters, GFLOPs and throughout in the image backbone and the rest of the network respectively: \textit{image backbone$\|$rest}.}
\label{tab:efficiency}
\end{table*}

%% file: tables/ablations_trajectory_singleVSmultiple.tex
\begin{table}[!ht]
    \centering
    \small
     \resizebox{\columnwidth}{!}{
    \begin{tabular}{l|cb}
    \toprule
    Trajectory representation & Multi-token &  Single-token \\
    \midrule
    PDMS & 83.9 & 90.0 \\
    \bottomrule
    \end{tabular}
    }
    \caption{\textbf{Single- vs multi-token trajectory.} 
    Quantitative comparison on \texttt{navval}.}
    \label{tab:ablations_trajectory_singleVSmultiple}
\end{table}

%% file: tables/navsim_v1.tex
\begin{table}[ht]
    \centering
    \small
    \setlength{\tabcolsep}{2pt}
    \resizebox{\columnwidth}{!}{
    \begin{tabular}{@{}l@{}r@{~}l|c|ccccc|c@{}}
    \toprule
    Method & & & Mod. &NC & DAC & TTC & \!Comf.\!\!& EP & PDMS \\
    \midrule
    \rowcolor{black!10}
    PDM‑Closed & \cite{dauner2023pdm}               & \scriptsize{PMLR'23} & - & 94.6 & 99.8 & 89.9 & 86.9 & 99.9 & 89.1 \\ 
    \rowcolor{black!10}
    Human driver& \cite{dauner2024navsim}           & \scriptsize{NeurIPS'24} & - & 100 & 100&  100 & 99.9 & 87.5 & 94.8 \\ 

    \midrule 
    \rowcolor{black!10}
    \multicolumn{10}{l}{\emph{Test-time training.}} \\
    \rowcolor{black!10}
    Centaur & \cite{sima2025centaur}                & \scriptsize{arXiv'25} & C & 99.5 & 98.9 & 98.0 & 100  & 85.9 & 92.6 \\ 

    \midrule
    \rowcolor{black!10}
    \multicolumn{10}{l}{\emph{Best-of-N scores.}} \\
    \rowcolor{black!10}
    AutoVLA  & \cite{zhou2025autovla} &  \scriptsize{NeurIPS'25} & C &  99.1 & 97.1 & 97.1 & 100.0 & 87.6 & 92.1 \\
    \rowcolor{black!10}
    DriveVLA-W0 & \cite{li2025drivevla} & \scriptsize{arXiv'25} & C & 99.3 & 97.4 & 97.0 & 99.9 & 88.3 & 93.0 \\   
    \rowcolor{black!10}
    TransDiffuser & \cite{jiang2025transdiffuser} & \scriptsize{arXiv'25} & C+L & 99.4 & 96.5 & 97.8 & 99.4 & 94.1 & 94.9 \\ 

    \midrule
    \rowcolor{black!10}
    \multicolumn{10}{l}{\emph{Ensemble methods.}} \\
    \rowcolor{black!10}
    Hydra-MDP-C \tiny(V2-99)\hspace{+2mm} &  \cite{li2024hydra} &\scriptsize{arXiv'25} & C+L &  98.7 & 98.2 & 95.0 & 100  & 86.5 & 91.0 \\ 

    \midrule
    \rowcolor{black!10}
    \multicolumn{10}{l}{\emph{Methods using  10× more training data.}} \\
    \rowcolor{black!10}
    RAP-DINO$^{\dagger}$  & \cite{feng2025rap}                   & \scriptsize{arXiv'25} & C & 99.1 & 98.9 & 96.7 & 100 & 90.3 & 93.8 \\ 

    \midrule
    \multicolumn{10}{l}{\emph{Multi-modal methods.}} \\
    
    TransFuser & \cite{chitta2022transfuser} & \scriptsize{TPAMI'22}& C+L & 97.7 & 92.8 & 92.8 & 100 & 79.2 & 84.0 \\ 
    DistillDrive & \cite{yu2025distilldrive} & \scriptsize{arXiv'25} & C+L & 98.1 & 94.6 & 93.6 & 100 & 81.0 & 86.2 \\ 
    TrajHF (EM) & \cite{li2025learningpersonalizeddrivingstyles} & \scriptsize{arXiv'25} & C+L &	96.6 & 96.6 & 92.1 & 100 & 84.5 & 87.6 \\ 
    DiffusionDrive & \cite{liao2025diffusiondrive} & \scriptsize{CVPR'25} & C+L & 98.2 & 96.2 & 94.7 & 100 & 82.2 & 88.1 \\ 
    WOTE & \cite{li2025end} & \scriptsize{ICCV'25} & C+L & 98.5 & 96.8 & 94.9 & 99.9 & 81.9 & 88.3 \\ 
    Hydra-MDP \tiny(V2-99) & \cite{li2024hydra} & \scriptsize{arXiv'24} & C+L &  98.0 & 97.8 & 93.9 & 100 & 86.5 & 90.3 \\ 
    GoalFlow \tiny(V2-99) & \cite{xing2025goalflow} & \scriptsize{CVPR'25} & C+L & 98.4 & 98.3 & 94.6 & 100 & 85.0 & 90.3 \\ 
    ResAD \tiny(V2-99) & \cite{zheng2025resad} & \scriptsize{arXiv'25} & C+L & 98.9 & 97.8 & 94.9 & 100 & 87.0 & 90.6 \\ 
    SeerDrive \tiny(V2-99) & \cite{zhang2025future} & \scriptsize{NeurIPS'25} & C+L &  98.8 & 98.6 & 95.8 & 100 & 84.2 & 90.7 \\ 
    
    \midrule
    \multicolumn{10}{l}{\emph{Camera-only methods.}} \\
    Ego‑stat. MLP & \cite{dauner2024navsim}        & \scriptsize{NeurIPS'24} & C & 93.0 & 77.3 & 83.6 & 100 & 62.8 & 65.6 \\ 
    UniVLA  & \cite{wang2025unified}                & \scriptsize{arXiv'25} & C & 96.9 & 91.1 & 91.7 & 96.7 & 76.8 & 81.7 \\ 
    DrivingGPT & \cite{chen2024drivinggpt}          & \scriptsize{ICCV'24} & C & 98.9 & 90.7 & 94.9 & 95.6 & 79.7 & 82.4 \\ 
    UniAD & \cite{hu2023uniad}                   & \scriptsize{CVPR'23} & C & 97.8 & 91.9 & 92.9 & 100  & 78.8 & 83.4 \\ 
    LTF & \cite{chitta2022transfuser}               & \scriptsize{TPAMI'22} & C & 97.4 & 92.8 & 92.4 & 100  & 79.0 & 83.8 \\ 
    PARA‑Drive & \cite{paradrive}               & \scriptsize{CVPR'24} & C & 97.9 & 92.4 & 93.0 & 99.8 & 79.3 & 84.0 \\ 
    DriveX-S & \cite{shi2025drivex}                 & \scriptsize{ICCV'25} & C & 97.5 & 94.0 & 93.0 & 100  & 79.7 & 84.5 \\ 
    World4Drive & \cite{zheng2025world4drive}       & \scriptsize{ICCV'25} & C & 97.4 & 94.3 & 92.8 & 100  & 79.9 & 85.1 \\ 
    DRAMA & \cite{yuan2024drama}                    & \scriptsize{ISRR'24} & C & 98.0 & 93.1 & 94.8 & 100  & 80.1 & 85.5 \\ 
    VAD-v2 & \cite{chen2024vadv2}                   & \scriptsize{arXiv'24} & C & 98.1 & 94.8 & 94.3 & 100 & 80.6 & 86.2 \\ 
    PRIX & \cite{wozniak2025prix}                   & \scriptsize{arXiv'25} & C & 98.1 & 96.3 & 94.1 & 100  & 82.3 & 87.8 \\ 
    DiffusionDrive & \cite{liao2025diffusiondrive} & \scriptsize{CVPR'25} & C & 98.2 & 96.2 & 94.7 & 100  & 82.2 & 88.1 \\ 
    DIVER & \cite{song2025breaking}                 & \scriptsize{arXiv'25} & C & 98.5 & 96.5 & 94.9 & 100  & 82.6 & 88.3 \\ 
    AutoVLA & \cite{zhou2025autovla}                & \scriptsize{NeurIPS'25} & C & 98.4 & 95.6 & 98.0 & 99.9 & 81.9 & 89.1 \\ 
    DriveVLA-W0 & \cite{li2025drivevla}             & \scriptsize{arXiv'25} & C & 98.7 & 99.1 & 95.3 & 99.3 & 83.3 & 90.2 \\ 
    ReCogDrive & \cite{li2025recogdrive}            & \scriptsize{arXiv'25} & C & 97.9 & 97.3 & 94.9 & 100  & 87.3 & 90.8 \\ 
    Hydra-MDP++   & \cite{li2025hydramdppp}         & \scriptsize{arXiv'25} & C & 98.6 & 98.6 & 95.1 & 100  & 85.7 & 91.0 \\ 
    iPad & \cite{guo2025ipad}                       & \scriptsize{arXiv'25} & C & 98.6 & 98.3 & 94.9 & 100  & 88.0 & 91.7 \\ 
    DriveSuprim  & \cite{yao2025drivesuprim}         & \scriptsize{arXiv'25} & C &98.6 & 98.6 & 95.5 & 100 & 91.3 & 93.5 \\
    \rowcolor{blue!15}
    \rowcolor{blue!15}
    \method{} \tiny{(train)}   && & C& 98.9 & 98.3 & 96.2 & 100 & 89.1 &  93.1      \\
    \rowcolor{blue!15}
    \method{} \tiny{(trainval)}   && & C & 99.0 & 98.9 & 96.7 & 100 & 90.0 &  93.7      \\
    \rowcolor{blue!15}
    \method{} \tiny{(+65k SimScale data)}   && & C& 99.1 & 99.0 & 96.9 & 100 & 90.3 & 94.0      \\
    \rowcolor{blue!15}
    \method{} \tiny{(+134k SimScale data) \hspace{+1mm}}   && & C& 99.1 & 99.2 & 96.9 & 100 & 91.6 & \bf 94.6      \\

    \bottomrule
    \multicolumn{9}{@{}l@{}}{$^\dagger$: \scriptsize{RAP~\cite{feng2025rap} is trained on a dataset that is 10$\times$ larger than navtrain (the default training set).}}
    \end{tabular}
    }
    \caption{\textbf{NAVSIM-v1 scores.} Full comparison to existing methods, possibly with different modalities (Mod.), on the NAVSIM-v1 benchmark on test set (\texttt{navtest}). `C' refers to camera, and `L' to LiDAR.
    }
    \label{tab:benchmark_navsim_v1}
\end{table}

%% file: tables/navsim_v2.tex
\begin{table*}[t]
    \centering
    \small
    \setlength{\tabcolsep}{2pt}
    \resizebox{\linewidth}{!}{
    \begin{tabular}{@{}l@{}r|ccccccccc|ccccccccc|c@{}}
    \toprule
            & & \multicolumn{9}{c}{Stage 1} & \multicolumn{9}{c}{Stage 2} & \\
    Method & &NC & DAC & DDC & TLC & EP & TTC & LK & HC & EC & NC & DAC & DDC & TLC & EP & TTC & LK & HC & EC & EPDMS \\
    \midrule
    \rowcolor{black!10}
    \multicolumn{21}{@{}l}{\emph{Results before official metric bug fixing.}}\\
    \rowcolor{black!10}
    PDM-Closed & \cite{dauner2023pdm} & 
    94.4 & 98.8 & 100 & 99.5 & 100 & 93.5 & 99.3 & 97.7 & 36.0 & 
    88.1 & 90.6 & 96.3 & 98.5 & 100 & 83.1 & 73.7 & 91.5 & 25.4 & 51.3 \\ 
    \rowcolor{black!10}
    Const. Vel. & \cite{dauner2024navsim} & 
    88.8 & 42.8 & 70.6 & 99.3 & 77.5 & 87.3 & 78.6 & 97.1 & 60.4 & 
    83.2 & 59.1 & 76.5 & 98.0 & 71.3 & 81.1 & 47.9 & 97.1 & 61.9 & 10.9  \\ 
    \rowcolor{black!10}
    Ego Hist. MLP & \cite{dauner2024navsim} & 
    93.2 & 55.7 & 86.6 & 99.3 & 81.2 & 92.2 & 83.5 & 97.5 & 77.7 & 
    77.2 & 51.9 & 74.4 & 98.2 & 77.1 & 75.0 & 40.8 & 97.8 & 79.8 & 12.7 \\ 
    \rowcolor{black!10}
    LTF & \cite{chitta2022transfuser} & 
    96.2 & 79.6 & 99.1 & 99.6 & 84.1 & 95.1 & 94.2 & 97.6 & 79.1 & 
    77.8 & 70.2 & 84.3 & 98.1 & 85.1 & 75.7 & 45.4 & 95.8 & 76.0 & 23.1 \\

    \rowcolor{black!10}
    RAP-DINO \tiny{(ViT-H)} $^\dagger$ & \cite{feng2025rap} &
    97.1 & 94.4 & 98.8 & 99.8 & 83.9 & 96.9 & 94.7 & 96.4 & 66.2 & 
    83.2 & 83.9 & 87.4 & 98.0 & 86.9 & 80.4 & 52.3 & 95.2 & 52.4 & 36.9 \\
    
    \rowcolor{black!10}
    GTRS-D \tiny{(V2-99)} & \cite{li2025generalized} &
    98.7 &91.4 &95.8 &89.2 &99.4 &94.4 &99.3 &98.8&72.8 &69.5 &
    98.7 &90.1 &95.1& 54.6& 96.9 &94.1 &40.4 & 49.7 & 41.7\\
    

    \rowcolor{black!10}
    GTRS-A \tiny{(V2-99)} & \cite{li2025generalized} &
     98.9 &  87.9 &95.1& 88.8 &99.2 &89.6& 99.6& 98.8& 76.1& 
     80.3& 99.1 & 86.0& 94.7& 53.5& 97.6& 97.1 &54.2& 56.1 & 42.1 \\
    
    \rowcolor{blue!5}
    GTRS-DrivoR \tiny{(ViT-S)}$^{*}$ & &
    98.0 & 95.8 & 99.7 & 99.3 & 72.9 & 98.2 & 95.6 & 96.9 & 51.6 & 
    91.6 & 86.7 & 90.2 & 98.8 & 73.2 & 88.9 & 51.9 & 94.9 & 46.4 & 42.3 \\
    
    \rowcolor{black!10}
    GTRS-D \tiny{EVA-ViT-L} & \cite{li2025generalized} 
    & 97.6 & 95.8 & 99.8 & 99.0 & 77.2 & 97.8 & 95.3 & 97.3 & 46.7 & 
    91.9 & 91.3 & 92.7 & 99.0 & 72.7 & 90.4 & 53.8 & 94.1 & 41.6 & 43.4\\
     
    \rowcolor{black!10}
    GTRS-A \tiny{(ViT-L)} & \cite{li2025generalized} & 
    98.7 & 98.0 & 99.1 & 99.8 & 75.9 & 98.7 & 94.7 & 97.6 & 49.8 &
    89.5 & 89.6 & 92.9 & 98.5 & 78.9 & 86.4 & 55.3 & 96.5 & 52.7 & 44.7 \\

    \rowcolor{black!10}
    DriveSuprim \tiny{(EVA-ViT-L)} \hspace{+1mm} & \cite{yao2025drivesuprim} &
    98.7 & 98.0 & 99.1 & 99.8 & 75.9 & 98.7 & 94.7 & 97.6 & 49.8 &
    89.5 & 89.6 & 92.9 & 98.5 & 78.9 & 86.4 & 55.3 & 96.5 & 52.7 & 44.7 \\ 

    \rowcolor{black!10}
    GTRS-D \tiny{(ViT-L)} & \cite{li2025generalized} &
    98.9 & 98.2 & 99.8 & 99.6 & 73.9 & 98.9 & 95.3 & 97.3 & 40.0 &
    91.5 & 90.8 & 94.7 & 98.5 & 70.8 & 90.1 & 55.4 & 97.2 & 54.2 & 45.3 \\
    \rowcolor{blue!15}
    \method  \tiny{(ViT-S)}  & & 98.8 & 95.1& 98.9& 100& 72.6&98.7&94.0&97.6&73.3&90.2&88.4&91.9&98.6&69.8&88.0&50.1&98.5&76.2&45.3\\
    \rowcolor{black!10}
    ZTRS \tiny{(V2-99)} & \cite{li2025ztrs} &
    98.9 & 97.6 & 100.0 & 100.0 & 66.7 & 98.9 & 96.2 & 96.7 & 44.0 &
    91.1 & 90.4 & 95.8 & 99.0 & 63.6 & 89.8 & 60.4 & 97.6 & 66.1 & 45.5 \\

    \midrule
    \multicolumn{21}{@{}l}{\emph{Results after official metric bug fixing.}}\\
    RAP-DINO \tiny{(ViT-H)}$^\dagger$ & \cite{feng2025rap}  & 97.1	&94.4	&98.8&	99.8&	83.9&	96.9	&94.7	&96.4	&66.2&	83.2&83.9&	87.4	&98.0	&86.9	&80.4	&52.3	&95.2&	52.4 &39.6\\
    GTRS-D \tiny{(V2-99)} & \cite{li2025generalized} &
     98.9  &96.2  &99.4 & 99.3 & 72.9 & 98.9 &95.1  &96.9 & 39.1 &
    91.2 & 89.4 &94.4 & 98.8 &69.5  &  90.0 & 54.3 &94.0  & 48.7& 45.0\\
    GTRS-A \tiny{(V2-99)} & \cite{li2025generalized} &
      98.9  &95.1& 99.1& 99.6 & 76.2 & 99.1  & 94.9 &97.6& 54.2 &88.1& 
      88.8 &89.3 &  98.9& 98.9&  85.9& 53.7&  96.8&56.9& 45.4\\

    \rowcolor{blue!5}
    GTRS-DrivoR \tiny{(ViT-S)}$^{*}$ &  &  98.0    &  95.8   &   99.7      &99.3       &72.9    &98.2     & 95.6     & 96.9   &   51.6    & 91.6    &  86.7  & 90.2   &98.8 &73.2     & 88.9     &51.9    & 94.9       &46.4 &45.8\\

    ZTRS \tiny{(V2-99)} & \cite{li2025ztrs} &
    98.9 & 97.6 &100&	100	&66.7&	98.9&	96.2&	96.7&	44.0	&
    91.1 & 90.4&	95.8&	99.0	&63.6&	89.8 &60.4&	97.6&	66.1& 48.1	\\
    \rowcolor{blue!15}

    \method{} \tiny{(ViT-S)} & &98.8& 95.1& 98.9& 100& 72.6&98.7&94.0&97.6&73.3&90.2&88.4&91.9&98.6&70.0&88.0&50.1&98.5&76.2&  48.3\\

    \rowcolor{blue!15}
    \method{} \tiny{(+65k SimScale data, ViT-S)} & &98.9& 97.3& 99.2& 99.6& 77.7&99.1&95.3&97.6&68.4&92.3&92.2&97.0&99.0&72.1&90.3&56.3&97.1&38.8&  52.3\\

    SimScale \tiny{(+185k SimScale data, V2-99)}&~\cite{tian2025simscale} & 99.6	& 99.16	& 99.9	& 100	& 69.6	& 99.6	& 95.8	& 95.6	& 28.4 & 94.5	&94.2	&95.8	&99.2&	75.8	&92.8	&60.1	&96.1	&43.2 & 53.2 \\

    \rowcolor{blue!15}
    \method{} \tiny{(+134k SimScale data, ViT-S)} & &99.1& 98.2& 99.3& 99.8& 75.4&98.7&94.9&97.6&70.2&92.3&91.6&97.3&99.1&75.7&90.6&56.1&98.4&44.7& \bf 54.6\\

    \bottomrule
    \multicolumn{19}{@{}l@{}}{$^\dagger$: \scriptsize{RAP~\cite{feng2025rap} is trained on a dataset that is 10$\times$ larger than navtrain (the default training set).}}\\
    \multicolumn{19}{@{}l@{}}{$^*$: \scriptsize{same ViT-S backbone with \method registers, the prediction and scoring heads remain the same as in GTRS.}}
    \end{tabular}
    }
    \vspace{-7pt}
    \caption{\textbf{NAVSIM-v2 \texttt{navhard-two-stage}}. Full comparison to other methods on the NAVSIM-v2 benchmark test set using the EPDMS. \textit{GTRS-A} refers to \textit{GTRS-Aug} and \textit{GTRS-D} for \textit{GTRS-Dense}.
    }
    \label{tab:benchmark_navsim_v2_small_supp}
\end{table*}